\documentclass[journal]{IEEEtran}
\usepackage[table]{xcolor}
\usepackage[numbers]{natbib}
\usepackage{amsmath,amssymb}
\usepackage{graphicx}
\usepackage{float}
\usepackage{tabularx,multirow}
\usepackage{booktabs}
\usepackage{hyperref}
\usepackage{makecell}
\usepackage{arydshln}
\usepackage{pifont}
\usepackage{orcidlink}
\usepackage{enumerate}
\usepackage{stfloats}
\usepackage{colortbl}

\definecolor{darkgreen}{HTML}{338309}
\definecolor{darkred}{HTML}{ff0066}
\definecolor{codegreen}{HTML}{527A6A}
\definecolor{codepink}{HTML}{DB5CA3}
\definecolor{darkblue}{HTML}{0066cc}
\definecolor{skyblue}{HTML}{96C5F8}
\definecolor{grassgreen}{HTML}{A0C97D}
\definecolor{orange}{HTML}{FFBE93}
\definecolor{gold}{HTML}{F5DF6C}
\definecolor{pink}{HTML}{CD9B9B}
\definecolor{textgreen}{HTML}{00b300}

\hypersetup{
colorlinks=true,
linkcolor=black,
citecolor=black
}

\begin{document}
\title{Exploring Generalizable Pre-training for Real-world Change Detection via Geometric Estimation}

\author{
	Yitao Zhao \hspace{-2mm}$^{~\orcidlink{0000-0001-6605-1757}}$,
        Sen Lei \hspace{-2mm}$^{~\orcidlink{0000-0002-8010-3282}}$,        
        Nanqing Liu \hspace{-2mm}$^{~\orcidlink{0000-0001-7564-4896}}$,
        Heng-Chao Li$^{*}$ \hspace{-2mm}$^{~\orcidlink{0000-0002-9735-570X}}$,
        Turgay Celik \hspace{-2mm}$^{~\orcidlink{0000-0001-6925-6010}}$,   
        and Qing Zhu \hspace{-2mm}$^{~\orcidlink{0000-0002-0485-4965}}$
        
	\IEEEcompsocitemizethanks{

		\IEEEcompsocthanksitem This work was supported in part by the National Natural Science Foundation of China under Grant 62271418, and in part by the Natural Science Foundation of Sichuan Province under Grant 2023NSFSC0030 and 2025ZNSFSC1154, in part by the Postdoctoral Fellowship Program and China Postdoctoral Science Foundation under Grant Number BX20240291, and in part by the Fundamental Research Funds for the Central Universities under Grant 2682025CX033. (Corresponding author: Heng-Chao Li)

		Yitao Zhao, Sen Lei, Nanqing Liu, and Heng-Chao Li are with the School of Information Science and Technology, Southwest Jiaotong University, Chengdu 611756, China (e-mail: ytzhao@my.swjtu.edu.cn; lansing163@163.com; lihengchao78@163.com). Turgay Celik is with the School of Information Science and Technology, Southwest Jiaotong University, Chengdu 611756, China, also with the School of Electrical and Information Engineering, University of the Witwatersrand, Johannesburg 2000, South Africa, and also with the Faculty of Engineering and Science, University of Agder, 4604 Kristiansand, Norway (e-mail: celikturgay@gmail.com). Qing Zhu is with the Faculty of Geosciences and Engineering, Southwest Jiaotong University, Chengdu 611756, China. (email: zhuq66@263.net)}
}

\IEEEtitleabstractindextext{
        \vspace{-0.5em}
	\begin{abstract}
        As an essential procedure in earth observation system, change detection (CD) aims to reveal the spatial-temporal evolution of the observation regions. A key prerequisite for existing change detection algorithms is aligned geo-references between multi-temporal images by fine-grained registration. However, in the majority of real-world scenarios, a prior manual registration is required between the original images, which significantly increases the complexity of the CD workflow. In this paper, we proposed a self-supervision motivated CD framework with geometric estimation, called ``MatchCD''. Specifically, the proposed MatchCD framework utilizes the zero-shot capability to optimize the encoder with self-supervised contrastive representation, which is reused in the downstream image registration and change detection to simultaneously handle the bi-temporal unalignment and object change issues. Moreover, unlike the conventional change detection requiring segmenting the full-frame image into small patches, our MatchCD framework can directly process the original large-scale image (e.g., $6K*4K$ resolutions) with promising performance. The performance in multiple complex scenarios with significant geometric distortion demonstrates the effectiveness of our proposed framework. 
	\end{abstract}

	}

\maketitle
\IEEEdisplaynontitleabstractindextext
\IEEEpeerreviewmaketitle

\section{Introduction}

\begin{figure}[!htp]
	\centering
	\includegraphics[width=0.83\linewidth]{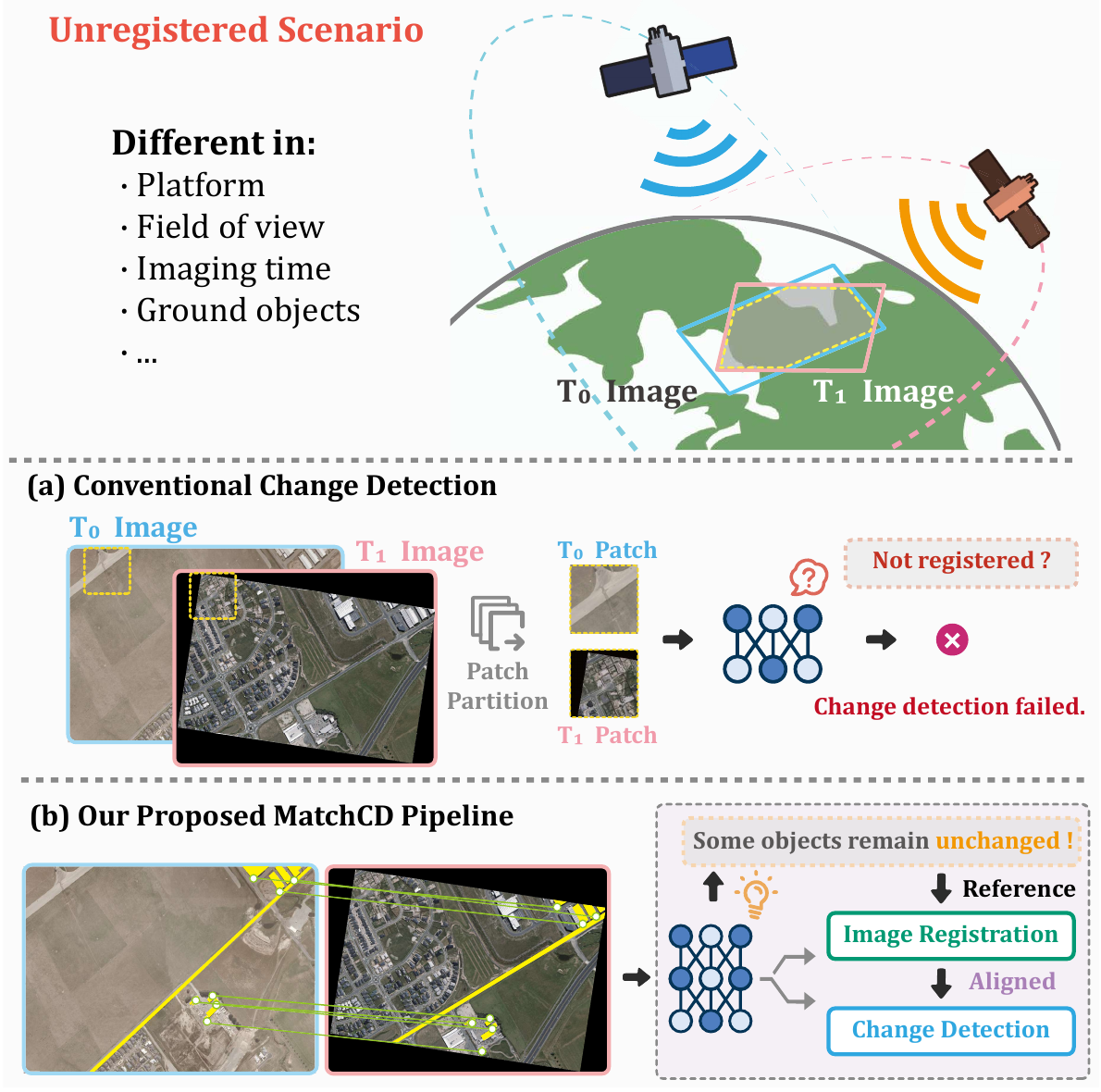}
	\caption{\textbf{Conventional CD methods vs. Our MatchCD.} Illustration of the difference between conventional CD methods and our MatchCD in unregistered scenario.}
	\label{fig:motivation}
\end{figure}

\IEEEPARstart{R}{mote} sensing change detection (RS CD) aims at extracting the changed regions by feature extraction and joint analysis in multi-temporal image sequences, which is an indispensable component of the RS image processing system \cite{liu2019review, huang2020automatic}. The accurate localization of change targets is of great significance in the fields of land use and land cover (LULC), urban planning and disaster assessment \cite{shafique2022deep, hao2023review}. Benefiting from the rapid advancement of RS imaging technology, massive amount of RS data are acquired by various remote sensing platforms with higher updating frequency, which provides extensive data support for earth observation system. Based on this, the performance of various RS techniques is also complementarily enhanced from the multi-source data \cite{chen2023fourier, zhang2023self, wang2023ssl4eo, lei2024exploring, lin2023ssmae}.

Conventional CD methods focus on revealing the change patterns between the multi-temporal images by supervised or unsupervised optimization with architectures like CNN \cite{fang2021snunet, zhang2020feature, liu2023attention, cao2023full, gao2019sea}, Vision Transformer \cite{chen2021remote, saleh2024dam, wang2024global, li2024ida}, Vision Mamba \cite{zhang2024novel, chen2024changemamba, zhang2024cdmamba} or hybrid architectures \cite{zheng2024unifying, paranjape2024mamba, zhang2024dc}. Afterwards, the change masks are generated for the specified class of objects. Generally, the backbone is first exploited to separate the target information from the background in RS images, and then the bi-temporal target features are compared to calculate the changed regions.  

The continual evolution of model architectures facilitates the incremental growth of the performance of CD methods. Nevertheless, an important prerequisite for the vast majority of change detection research is that the bi-temporal images for comparison are accurately matched. Specifically, all existing public CD datasets are obtained by manual pre-processing (e.g., geometric alignment and segmentation), and the CD algorithms fail to consider the cost of such pre-processing. For example, the geometric distortion is neglected in some widely used datasets like \textit{LEVIR-CD} dataset \cite{chen2020spatial}, \textit{WHU-CD} dataset \cite{ji2019full}, and \textit{CDD} dataset \cite{lebedev2018change}. However, as the research horizon expands to real-world scenarios, RS images from different platforms may possess quite different coordinate systems. As in the case shown in Fig. \ref{fig:motivation}-(a),  given two large size aerial images (e.g., 6K*4K), conventional CD pipelines require pre-processing of patch segmentation according to the pixel coordinates, and then send the segmented patches into the model for feature extraction. However, in the global unaligned scenario shown in Fig. \ref{fig:motivation}-(a), the CD classifier receives the segmented bi-temporal patches with completely distinct contents, resulting in the eventual failure of changed region extraction. 

Available solutions to the aforementioned unregistered issue are relatively homogeneous. Typically, manual control point selection or fast image registration algorithms are first employed for global correction to eliminate the geometric distortions. Some researchers proposed the learning-base image registration methods, which overcome the requirement of manually designed descriptors for traditional algorithms \cite{sun2021loftr, sarlin2020superglue, lindenberger2023lightglue, edstedt2023dkm}. Then the segmented patches could be provided to the subsequent vision tasks. However, such method causes discontinuities in the workflow since the registration network needs to be additionally designed and trained \cite{zhao2024towards}. Furthermore, since lacking fixed format for annotation, the point-level labeling for the image registration tasks is laborious and challenging. 

In recent years, the RS community is undergoing a tremendous revolution due to the development of foundation models \cite{kirillov2023segment, radford2021learning, lei2021transformer}. A number of separate downstream tasks are composed into an integrated system with the robust features provided by the foundation models. The utilization of visual foundation models sufficiently improves the limitation of traditional RS tasks \cite{hong2024spectralgpt, liu2024pointsam, ma2024sam, chen2024rsprompter}. Moreover, additional prompt information such as text or geo-location inherited in multimodal foundation models can also provide complementary enhancement \cite{sun2022ringmo, dong2024changeclip, kuckreja2024geochat}. Motivated by this, we rethink the CD workflow towards real-world scenarios. Since image registration and change detection are tightly related, we can fully leverage the common patterns between both the tasks. Specifically, for the matching between dual-view scenarios in which change detection will be performed subsequently, we can first establish feature correlation based on the \textit{unchanged objects}, and then the changed regions could be compared and extracted in the aligned coordinate system. As shown in Fig. \ref{fig:motivation}-(b), since the two workflows share the same scenario, the zero-shot capability of the foundation models could be exploited to synchronize the enhancement of both tasks.

In this paper, we propose a generalizable pre-training framework for real-world change detection via geometric estimation. The proposed MatchCD framework is constructed in a two-stage approach, containing an instance-level contrastive pre-training stage, followed by a down-stream stage of full-scenario ``Registration-Detection''. In the pre-training stage, the zero-shot capability of foundation model is exploited for instance generation. During this stage, different views from single instance are prompted for clustering to capture the robustness of geometric transformation. Afterwards, a training-free image registration paradigm is established to perform full-scenario correction between the hierarchical features from the pre-trained encoder. It is worth noting that the geometric boundary of the common region in registered images will also be extracted and sent into the subsequent CD classifier to eliminate the anomaly caused by pixels outside the boundary.

The main contributions can be summarized as follows:
\begin{itemize}
    \item To the best of our knowledge, it is the first pioneering exploration of an end-to-end “registration-change detection” workflow. A generalized framework called MatchCD is proposed to directly detect building changes between large-scale bi-temporal images ($6K*4K$) with significant geometric distortions.

    \item An training-free registration paradigm is developed for pre-change and post-change RS images, in which the unchanged objects are leveraged for geometric estimation.
    
    \item We design a concise change detection classifier by utilizing the prior context provided by the pre-trained encoder and foundation model to enhance the change detection performance.

    \item A building change detection dataset called WarpCD is created and released for this research. Perspective variations are introduced during the data generation to simulate the impact of viewpoint change on the change detection task in realistic scenarios. 

\end{itemize}

The rest of this paper is organized as follows. Section \uppercase\expandafter{\romannumeral2} describes previous work related to this research. The detailed procedures of the proposed MatchCD framework are introduced in Section \uppercase\expandafter{\romannumeral3}. Afterwards, extensive experimental results and discussions of image registration and change detection in real-world scenarios are presented in Section \uppercase\expandafter{\romannumeral4} and Section \uppercase\expandafter{\romannumeral5}.  Finally, we present the summary and perspectives in Section \uppercase\expandafter{\romannumeral6}.

\section{Related Work}

\subsection{Learning-based Image Registration}
Image registration is an fundamental procedure in multi-view tasks like time series RS analysis, 3D reconstruction and SLAM. In the field of natural image processing, numerous image registration methods are proposed to overcome the distortion between different views. Among the traditional registration methods, SIFT algorithm acts as essential milestone which is still commonly utilized in many fields \cite{lowe2004distinctive}. 

Beyond the traditional methods, some researchers propose several learning-based image registration ones by leveraging the powerful computational capability of deep learning. DeTone et al. propose a self-supervised training framework to obtain multi-view correlation capability. The proposed SuperPoint framework introduces homographic adaptation for boosting the repeatability of interest point detection \cite{detone2018superpoint}. On the basis of previous research, the SuperGlue framework is designed with attentional graph neural network to calculate the optimal matching parameters \cite{sarlin2020superglue}. With the emergence of vision transformer, such novel architecture is introduced to image matching task. LoFTR algorithm exploits the self-attention and cross-attention in ViT to acquire coarse-to-fine feature descriptors for global matching \cite{sun2021loftr}. To overcome the computational cost of previous research, LightGlue is organized in a flexible structure with attention mechanism, achieving higher inference performance and efficiency \cite{lindenberger2023lightglue}. Benefiting from the huge amount of online video data, the GIM framework establishes a self-training paradigm to optimize the generalizable algorithm from sequence images. Several well-known registration algorithms are proven to be strengthened by such paradigm \cite{xuelun2024gim}.  

In contrast to natural images, there remain some inherent challenges in RS image registration tasks. Firstly, the image resolution and scale of data from different RS platforms may vary significantly. Furthermore, natural images usually possess clear features, while RS images may suffer from fuzzy texture details caused by weather, illumination variations, or topographical factors. In addition, the prevalence of construction and demolition of ground objects as well as the seasonal changes in RS images also increase the difficulty of feature matching. To address these concerns, some specified improvements have been proposed.

Chen et al. propose a learning-based registration method for multimodal RS images, leveraging the similarity of latent features to eliminate the heterogeneous domain gap \cite{chen2020geometric}. To overcome the shortcoming of traditional manually designed descriptor, MCGF combines the structure information with deep features to improve the matching performance \cite{zhou2022robust}. In addition, a few studies attempt to incorporate the matching operation into CD training to resist the minor offsets between bi-temporal samples \cite{chen2023rdet, zhou2024unified}.

Overall, the majority of aforementioned works only consider image registration as an individual task, without exploring the correlation with other downstream tasks. Moreover, the registration procedure is limited between two small patches rather than the full-scenario matching in real-world task. 

\subsection{Self-supervised Change Detection}
Conventional change detection require extensive manual labeling for supervised optimization, which leads to significant time and financial costs. Although some automatic labeling tools have been released with the assistance of visual foundation models, the manual interaction during the labeling still remain critical to ensure the labeling accuracy \cite{kirillov2023segment}. Therefore, some researchers shift the research direction to the self-supervised field. Typically, the self-supervised learning (SSL) establish the optimization stage with numerous of unlabeled homogeneous or heterogeneous data, in which the robust feature representation is obtained. Afterwards, the pre-trained feature could be transferred to downstream tasks for fine-tuning. Such learning paradigm has been widely adopted in NLP and CV communities to support the foundation model pre-training and various self-supervised based tasks \cite{he2020momentum, chen2020simple, chen2021exploring, he2022masked, hong2024spectralgpt}. 

The utilization of SSL in CD tasks is relatively limited. Following the trail of contrastive SSL, Jiang et al. propose a SimSiam-like framework for CD task pre-training with global-local contrast \cite{jiang2023self}. To eliminate the inter-domain variance between bi-temporal samples, SSLChange method introduces a domain adaption paradigm to the pretext task. Incorporated with the pre-trained SSLChange backbone, downstream CD backbones obtains significant enhancement under label-limited situation \cite{zhao2024sslchange}. In addition to contrastive SSL, masked image modeling also achieves promising performance in CD task. The multi-source data of RGB images and nDSM data are jointly exploited in RECM method to simultaneously perform contrastive and masked pre-training \cite{zhang2023self}. Under the time series scenario, COUD framework perform contrastive and masked SSL paradigms with the multi-temporal RS image sequence to acquire temporal leverages context \cite{zhao2024coud}. 

However, although the current researches on self-supervised CD significantly improve the generalization and reduce the dependency on labels compared to supervised methods, none of them consider the prior impact of geometric distortions in realistic scenarios.

\begin{figure*}[!htbp]
	\centering
        \scalebox{0.92}{
	\includegraphics[width=\linewidth]{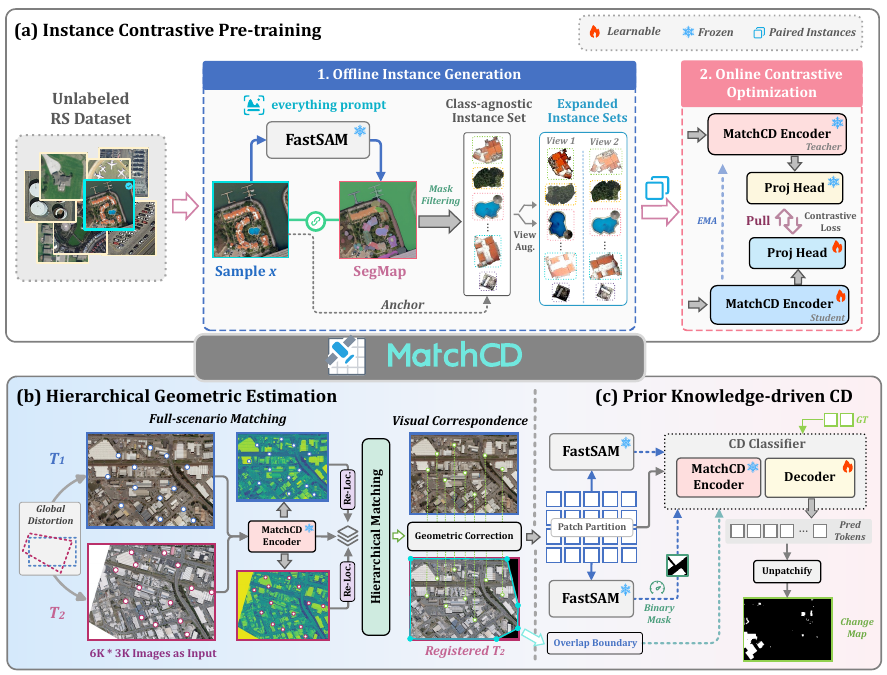}}
	\caption {Workflow of the proposed MatchCD framework. The whole MatchCD framework contains three main procedures: (a) Instance Contrastive Pre-training, (b) Hierarchical Geometric Estimation, (c) Prior Knowledge-driven CD. }
	\label{fig:workflow}
\end{figure*}

\section{Proposed Method}
The overview of MatchCD framework and the algorithm processes of key components will be elaborately described in this section. 

\subsection{Overview}

The overall workflow of MatchCD is illustrated in Figure \ref{fig:workflow}. The pre-training stage utilizes extensive instance samples generated by the FastSAM \cite{zhao2023fast} for contrastive clustering. In this stage, the label is not required for model optimization, and the generated instance samples are all class-agnostic. Afterwards, the geometric estimation is performed between two large scenarios with significant distortion. Besides the keypoints on the original image pairs, the candidate points from the hierarchical latent features are also extracted and merged. Following this, the pre-trained MatchCD encoder is exploited as the component of the CD classifier. In addition, the FastSAM is reused to generate prior knowledge of the input tokens. Finally, the pre-trained features and prior knowledge are jointly sent into the decoder for training.

\subsection{Instance Contrastive Pre-training}
\label{sec:pretrain}
In conventional workflow of self-supervised contrastive learning, the original sample is randomly cropped or rotated to obtain different views. Although such a method has achieved promising performance \cite{he2020momentum,chen2020simple,chen2021exploring}, the main object within the single sample may be cropped out after the transformation. In addition, the target distribution in RS images is generally denser than that it natural images, with more significant variations in object scales and geometric characteristics.

To fully leverage the object patterns on the RS images, we improve the aforementioned issues, proposing the instance contrastive pre-training with the zero-shot capability of visual foundation model. Here, we separate this stage into two procedures: (1) Offline Instance Generation, and (2) Online Contrastive Optimization. 

Given a single sample $x \in \mathbb{R}^{H \times W \times C}$ in the unlabeled RS image dataset, we first utilize the FastSAM \cite{zhao2023fast} to extract the objects from the sample $x$ and obtain the segmentation map with mask set $\mathbf{M}_x$. As illustrated in Fig. \ref{fig:workflow}-(a), for each segmented mask $m$ in $\mathbf{M}_x$, we multiple the mask $m$ with the original sample $x$ to calculate the corresponding instance sample:
\begin{align}
    \mathbf{M}_x &= FastSAM(x) \\
    \mathbf{I}_x &= \left\{ m \otimes x \mid m \in \mathbf{M}_x \right\}
    \label{instance}
\end{align}

After all the calculations completed by the above equations, all the instance samples are combined to get the instance set $\mathbf{I}_x$. During the above progress, we apply a mask filtering to exclude the masks with excessive or insufficient portion of pixel counts (e.g., small vehicles and background). Specifically, the mask with pixel count portion $p_m$ in the range of $10\% \leq p_m \leq 50\%$ is reserved for instance calculation. It is worth noting that the instance set here is class-agnostic in which the class information is not limited in building, since we believe that a wider class of instances will provide abundant patterns for the subsequent contrastive pre-training. 

Following this, the view augmentation is performed on each instance sample $\mathbf{I}_x^{i}$ in the class-agnostic instance set to generate various views with distinct style. Denoting the view augmentation operation as $\mathcal{T}$, the transferred views are calculated by:
\begin{equation}
	v_1=\mathcal{T}_{1}(\mathbf{I}_x^{i}), \quad v_2=\mathcal{T}_{2}(\mathbf{I}_x^{i})
\end{equation}
where $\mathcal{T}_{1}$ and $\mathcal{T}_{2}$ are the different view augmentation operations, and $v_1$ and $v_2$ represent the transferred views. In the implementation of MatchCD method, the random rotation and illumination variation is introduced into the view augmentation. Specifically, the random rotation employs a probability of 50\% to rotate the image by 90° or 180°, and the illumination variation simulation employs random color jittering to adjust the brightness, contrast, and saturation components to simulate samples acquired in different illumination scenarios. In the aforementioned offline pipeline, the expanded instance sets are obtained for the subsequent contrastive optimization. 

In the online contrastive optimization stage, the paired instance samples are sequentially sent to the established MatchCD encoder and projection head. The contrastive pre-training model employs the teacher-student architecture, in which the student encoder and projection head are optimized by the back propagation, and the teacher parts are updated with EMA algorithm. We denote the teacher and student encoders as $E_t$ and $E_s$, and the corresponding projection heads as $H_t$ and $H_s$. The output of the teacher model and student model are as follows:
\begin{align}
    p_{t}^{v1} = E_t(H_t(v_1)), \quad p_{t}^{v2} &= E_t(H_t(v_2)) \\
    p_{s}^{v1} = E_s(H_s(v_1)), \quad p_{s}^{v2} &= E_s(H_s(v_2)) 
\end{align}

Following the basic format of DINO loss \cite{caron2021emerging}, we establish the loss function of the pre-training stage. A clustering center $C$ is introduced and initialized as an all-zero matrix, then updated by $C \leftarrow m \cdot C + (1-m) \sum P_{t}^{i} $ during the training. $m$ is an momentum parameter set as 0.9. The cross-entropy loss is utilized as:
\begin{equation}
    D(x, y) = - \sigma((x-C) / \tau) \cdot log(\sigma(y / \tau))
\end{equation}
where, $\tau$ is the temperature parameter, and $\sigma$ is softmax function. Then, we perform the above loss function between the outputs of teacher model and student model:
\begin{align}
    \mathcal{L}_{Pre} = \frac{D(p_{t}^{v1}, p_{s}^{v2})}{2} + \frac{D(p_{t}^{v2}, p_{s}^{v1})}{2}
\end{align}

\begin{figure}[!htbp]
	\centering
	\includegraphics[width=1.0\linewidth]{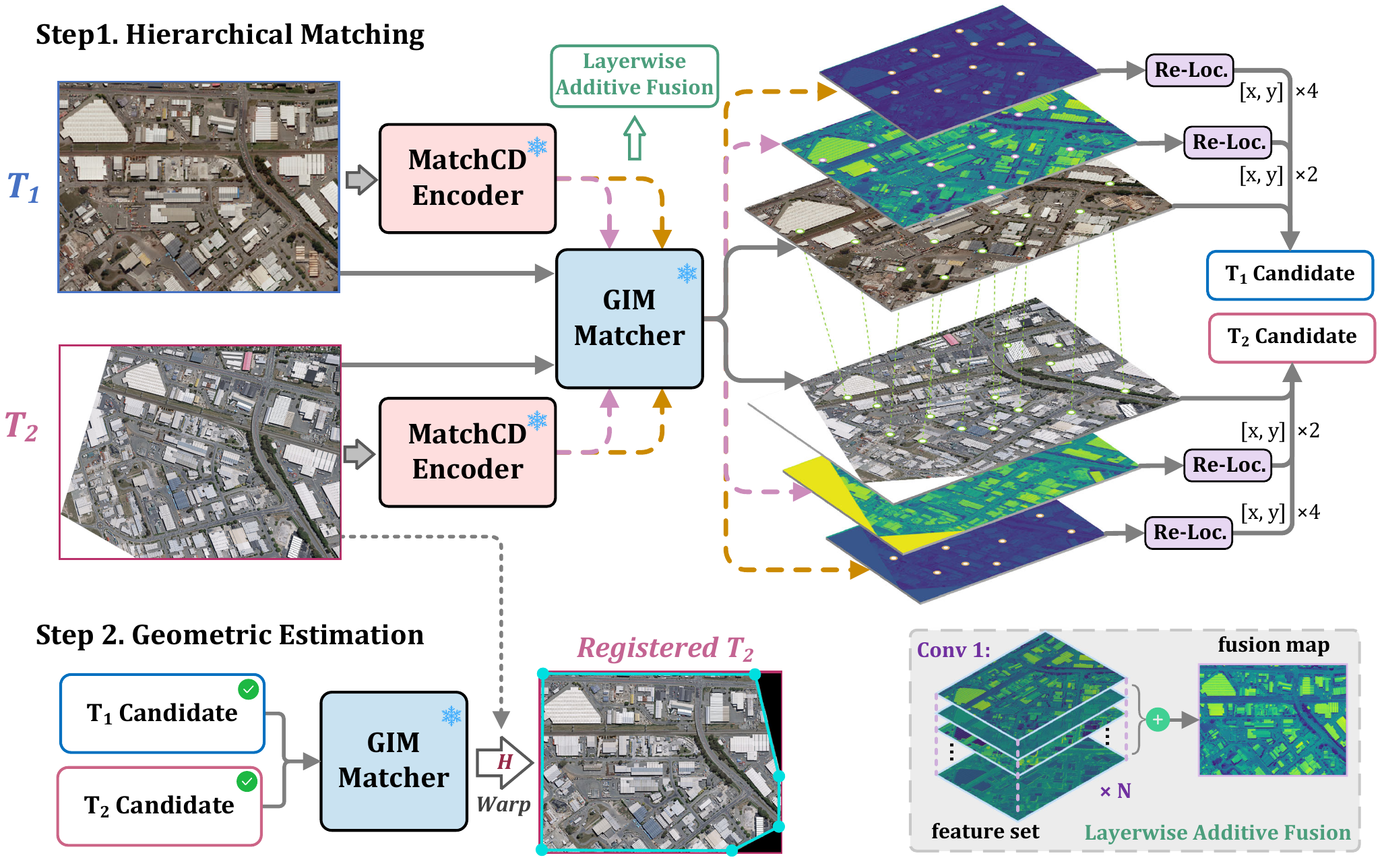}
	\caption{Illustration of the hierarchical geometric estimation for bi-temporal real-world scenario with significant distortion. }
	\label{fig:matching}
\end{figure}

Through the aforementioned optimization, the MatchCD encoder is prompted to capture the detailed features without any external supervision. In addition, since the view augmentation operation considers random transformation such as rotation and illumination variation, the encoder will possess the robustness to the bi-temporal geometric distortion.

\subsection{Hierarchical Geometric Estimation}
As shown in Fig. \ref{fig:workflow}-(b), the pre-trained MatchCD encoder is retained as the backbone of bi-temporal registration stage. We organize this stage with a training-free paradigm, in which the MatchCD encoder and the GIM Matcher \cite{xuelun2024gim} are all frozen to efficiently perform inference. 

\subsubsection{\textbf{Hierarchical Matching}}
In contrast to the existing CD research neglecting the matching issue, we expand the matching scenario from two small patches to a large real-world scenario with significant geometric distortion, where the conventional CD methods fail to be directly established. 

As illustrated in Fig. \ref{fig:matching}, given a pair of large bi-temporal images $I^{T_1}$ and $I^{T_2}$ (e.g., with resolution of $6K \times 3K$), the pre-trained MatchCD encoder is reused to extract latent features. The specific architecture of MatchCD encoder is a ResNet-50 pre-trained with the method proposed in section \ref{sec:pretrain}. We extract the shallow features generated by $\mathit{Conv 1}$ and $\mathit{Layer 1}$ from the encoder. Since the objects on the original RS images might be blended with the background, we argue that the latent features contain complementary semantic patterns with potential contribution to the subsequent matching. 

It is worth noting that the features of $\mathit{Conv 1}$ and $\mathit{Layer 1}$ are 64-dim and 256-dim, all the features from the same layer are gathered in a layerwise additive fusion manner to obtain the fusion map as shown in Fig. \ref{fig:matching}-\textit{Layerwise Additive Fusion}. 

Denoting the feature maps of image $I^{T_i}$ from $\mathit{Conv 1}$ and $\mathit{Layer 1}$ as $F_{C_1}^{T_i}$ and $F_{L_1}^{T_i}$, respectively. The hierarchical matching procedure utilizing GIM matcher is as follows:
\begin{align}
    \mathbf{P}_{Ori} &= \Theta_{GIM}(I^{T_1}, I^{T_2}) \\
    \mathbf{P}_{C_1} &= \Theta_{GIM}(F_{C_1}^{T_1}, F_{C_1}^{T_2}) \\
    \mathbf{P}_{L_1} &= \Theta_{GIM}(F_{L_1}^{T_1}, F_{L_1}^{T_2}) 
\end{align}
where, $\mathbf{P}_{Ori}$ is the keypoint set between $I^{T_1}$ and $I^{T_2}$. $\mathbf{P}_{C_1}$ and $\mathbf{P}_{L_1}$ represents the keypoint sets between the bi-temporal $\mathit{Conv1}$ and $\mathit{Layer1}$ feature maps.

Considering the downsampling ratios within the MatchCD encoder, we perform keypoint re-localization between the original image and generated feature maps. Specifically, given a detected keypoint $(x_{C_1}, y_{C_1})$ in set $\mathbf{P}_{C_1}$, the re-located keypoint set on the original image can be acquired by:
\begin{equation}
    \mathbf{P}_{Ori}^{\prime} = \{(x_{Ori}^{\prime}, y_{Ori}^{\prime})  \mid  x_{Ori}^{\prime}= 2x_{C_1}, y_{Ori}^{\prime} = 2y_{C_1} \}     
\end{equation}

Similarly, the re-located set from the $\mathit{Layer 1}$ feature to the original image can be represented as:
\begin{equation}
    \mathbf{P}_{Ori}^{\prime\prime} = \{(x_{Ori}^{\prime\prime}, y_{Ori}^{\prime\prime})  \mid  x_{Ori}^{\prime\prime}= 4x_{L_1}, y_{Ori}^{\prime\prime} = 4y_{L_1} \}  
\end{equation}

Accordingly, the final keypoint candidate set $\mathbf{KP}$ is the union of the hierarchical keypoint sets:
\begin{equation}
    \mathbf{KP} = \mathbf{P}_{Ori} \cup \mathbf{P}_{Ori}^{\prime} \cup  \mathbf{P}_{Ori}^{\prime\prime} 
\end{equation}

The specific elements in the final keypoint candidate set $\mathbf{KP}$ is as follows:
\begin{equation}
    \begin{split}
        \mathbf{KP} = \{ & [(x_{0}^{T1}, y_{0}^{T1}), (x_{1}^{T1}, y_{1}^{T1}),                        \dots, (x_{N}^{T1}, y_{N}^{T1})], \\
                         & [(x_{0}^{T2}, y_{0}^{T2}), (x_{1}^{T2}, y_{1}^{T2}),       \dots, (x_{N}^{T2}, y_{N}^{T2})] \}
    \end{split}
\end{equation}
where $N$ is the number of correlated point pairs. The first and second rows of $\mathbf{KP}$ can be represented as $\mathit{T_1}$ candidate and $\mathit{T_2}$ candidate shown in Fig. \ref{fig:matching}, respectively.

\subsubsection{\textbf{Geometric Estimation}}
Once the final keypoint candidate set $\mathbf{KP}$ is acquired, the homography matrix $H$ can be estimated to restore the mapping relationship between $I_{T_1}$ and $I_{T_2}$ coordinates. 

Denoting the $I^{T_1}$ and $I^{T_2}$ candidate set as $\mathbf{S}^{T_1}$ and $\mathbf{S}^{T_2}$. We set the homography matrix $H$ as a $3 \times 3$ matrix. For each paired coordinate from $\mathbf{S}^{T_1}$ and $\mathbf{S}^{T_2}$, the transformation between their homogeneous coordinates with $H$ is as follows:
\begin{equation}
    \begin{bmatrix}
    x_{i}^{T_1} \\ y_{i}^{T_1} \\ 1
    \end{bmatrix}
    = H \cdot
    \begin{bmatrix}
    x_{i}^{T_2} \\ y_{i}^{T_2} \\ 1
    \end{bmatrix}
\end{equation}

Once sufficient number of matching point pairs are available, the matrix $H$ can be solved by normalized $DLT$ algorithm combined with $RANSAC$ \cite{longuet1981computer, fischler1981random}. Afterwards, the image $I^{T_2}$ is transformed to the coordinate system of $I^{T_1}$ with the homography matrix $H$. 

In addition, to avoid the negative impact of invalid regions outside the transformed image boundary, the overlap detection is also perform on the warped image to get the boundary of public region. Specifically, we first construct two polygons with the four vertices of the $I^{T_1}$ and transformed $I^{T_2}$ respectively. Then the intersection coordinates of the two polygons are calculated. Finally, the warped image pairs with the overlap boundary are preserved as the basis of the subsequent change detection procedure. 

The experimental detail of image registration can be found in Section \ref{visual}.

\subsection{Prior Knowledge-driven Change Detection}

In this section, we reveal the concise process of implementing the pre-trained encoder and the prior knowledge from foundation model for downstream change detection task.

\begin{figure}[!htp]
    \centering
    \includegraphics[width=\linewidth]{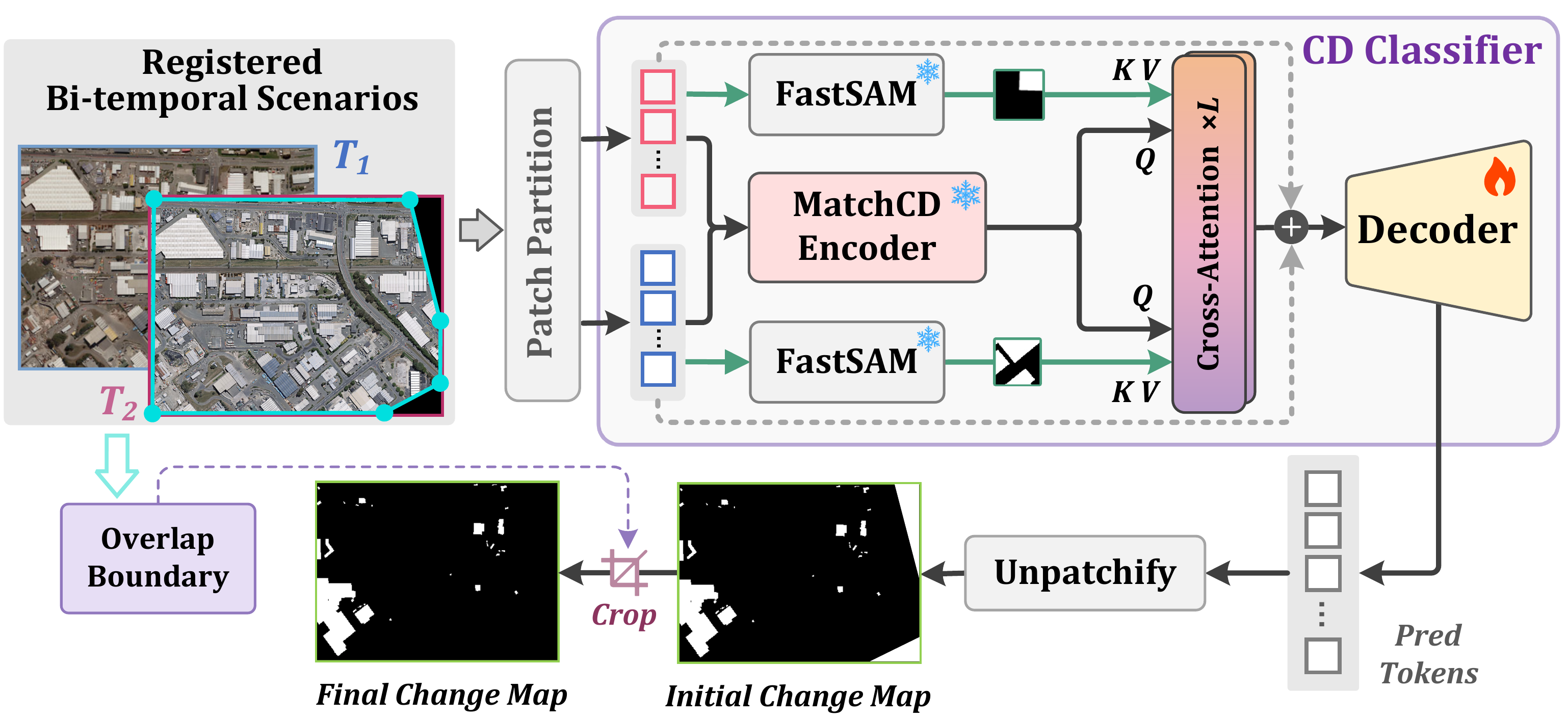}
    \caption{Illustration of downstream change detection workflow. }
    \label{fig:change_detection}
\end{figure}

For a pair of registered images, we first perform patch partition to segment the full scenarios into image patches. Denoting the acquired patches as $\mathbf{P}^{T_1}$ and $\mathbf{P}^{T_2}$, respectively. We first utilize the pre-trained MatchCD encoder to extract the latent feature. Meanwhile, the binary mask from FastSAM model is also calculated as the prior knowledge. The specific calculation process can be represented as follows:
\begin{align}
    e_{T1} &= \Theta_{MatchCD}(\mathbf{P}^{T_1}_i), \quad  bin_{T_1} = \Theta_{FastSAM}(\mathbf{P}^{T_1}_i) \\
    e_{T2} &= \Theta_{MatchCD}(\mathbf{P}^{T_2}_i), \quad  bin_{T_2} = \Theta_{FastSAM}(\mathbf{P}^{T_1}_i) 
\end{align}

To exploiting the guidance of multimodal prior prompts, the cross attention mechanism is introduced to merge the embeddings as follows:
\begin{align}
    token^{T_1}_i &= \mathbf{CA}(e_{T_1}, bin_{T_1}), \\
    token^{T_2}_i &= \mathbf{CA}(e_{T_2}, bin_{T_2})
\end{align}
where $token^{T_1}_i$ and $token^{T_2}_i$ are the tokens from the attention. 

Afterwards, a 5-layer U-Net++ decoder is attached to predict the mask tokens containing the changed regions. The predict tokens are stitched to restore the original resolution of change map. 
\begin{align}
    token^{T_1}_i &= \mathrm{concat}(\mathbf{P^{T_1}}_i, token^{T_1}_i) \\
    token^{T_2}_i &= \mathrm{concat}(\mathbf{P^{T_2}}_i, token^{T_2}_i) \\
    token^{pred}_i &= \mathbf{Dec}(token^{T_1}_i, token^{T_2}_i) \\
    M_{Init} &= \oplus token^{pred}_i
\end{align}
where $\mathbf{Dec}$ is the MatchCD decoder, and $token^{pred}_i$ is the $i\mbox{-}th$ prediction token from the decoder. $\oplus$ operation represents all the prediction tokens are concatenated to restore the dimensions of original large input image. 

Particularly, the overlap boundary is utilized in the post processing to crop the valid regions on the initial change map.
\begin{equation}
    M_{Final} = M_{Init} \odot Mask_{OB}
\end{equation}
where $\odot$ is the element-wise multiplication operation, and $Mask_{OB}$ is the mask image generated from the overlap boundary. 

During the network optimization, the prediction tokens are utilized to calculate the training loss along with the pixel-level labels. The optimization function is as follows:
\begin{align}
    \mathcal{L}_{CD}(y, \hat{y}) &= - \omega \times \mathit{\hat{y}} \cdot log(\mathit{y}) + (1-\mathit{\hat{y}}) \cdot (log(1-\mathit{y}))  \\
    \mathcal{L}_{MatchCD} &= \mathcal{L}_{CD}(token^{pred}, GT)
\end{align}
where $\omega$ is the weight parameter for the positive samples, and $GT$ are is the training labels.

The experimental detail of down-stream change detection can be found in Section \ref{quantitative} and Section \ref{visual}. 

\begin{figure*}[!htp]
    \centering
    \includegraphics[width=\linewidth]{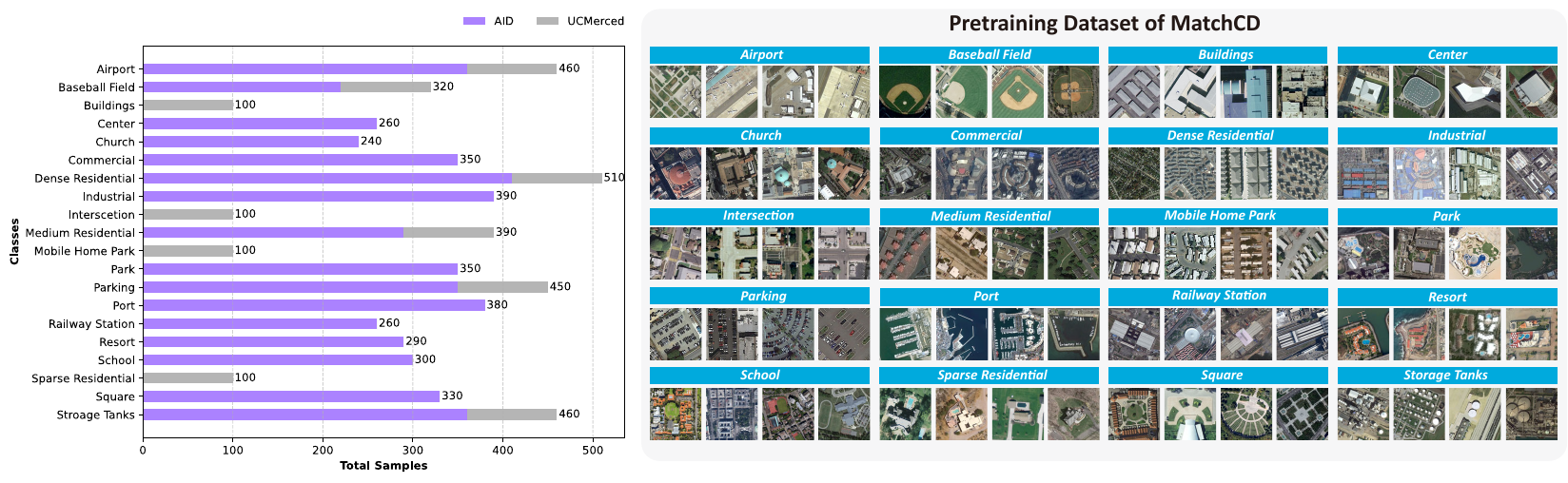}
    \caption{Illustration of the dataset utilized for MatchCD pre-training. The samples are selected from the \textit{AID} and \textit{UCMerced} dataset according to the class information related to building objects. } 
    \label{fig:pretrain}
\end{figure*}

\section{Experiments}
\subsection{Datasets}
\subsubsection{Pre-training Datasets} We use two large RS images datasets: \textit{AID} \cite{xia2017aid} and \textit{UCMerced} \cite{yang2010bag} for MatchCD pre-training. As shown in Fig. \ref{fig:pretrain}, considering the main detecting target of our research is building, we perform manually sample selection within the above two datasets according to the categories information provided by them. Specifically, we chose the samples with the categories which contain building objects to construct the pre-training dataset. A total of 20 categories are covered by the constructed pre-training dataset. The sample number of each category is listed in the histogram, and the visualizations of typical samples from category are provided on the right side of Fig. \ref{fig:pretrain}. All the selected samples are merged into an independent dataset. In the pre-training stage, the instance segmentation with FastSAM is performed on this dataset and generates over 51,000 instance samples for the contrastive optimization. 

\subsubsection{Downstream Datasets}

\begin{figure*}[!htbp]
    \centering
    \includegraphics[width=0.82\linewidth, height=0.6\linewidth]{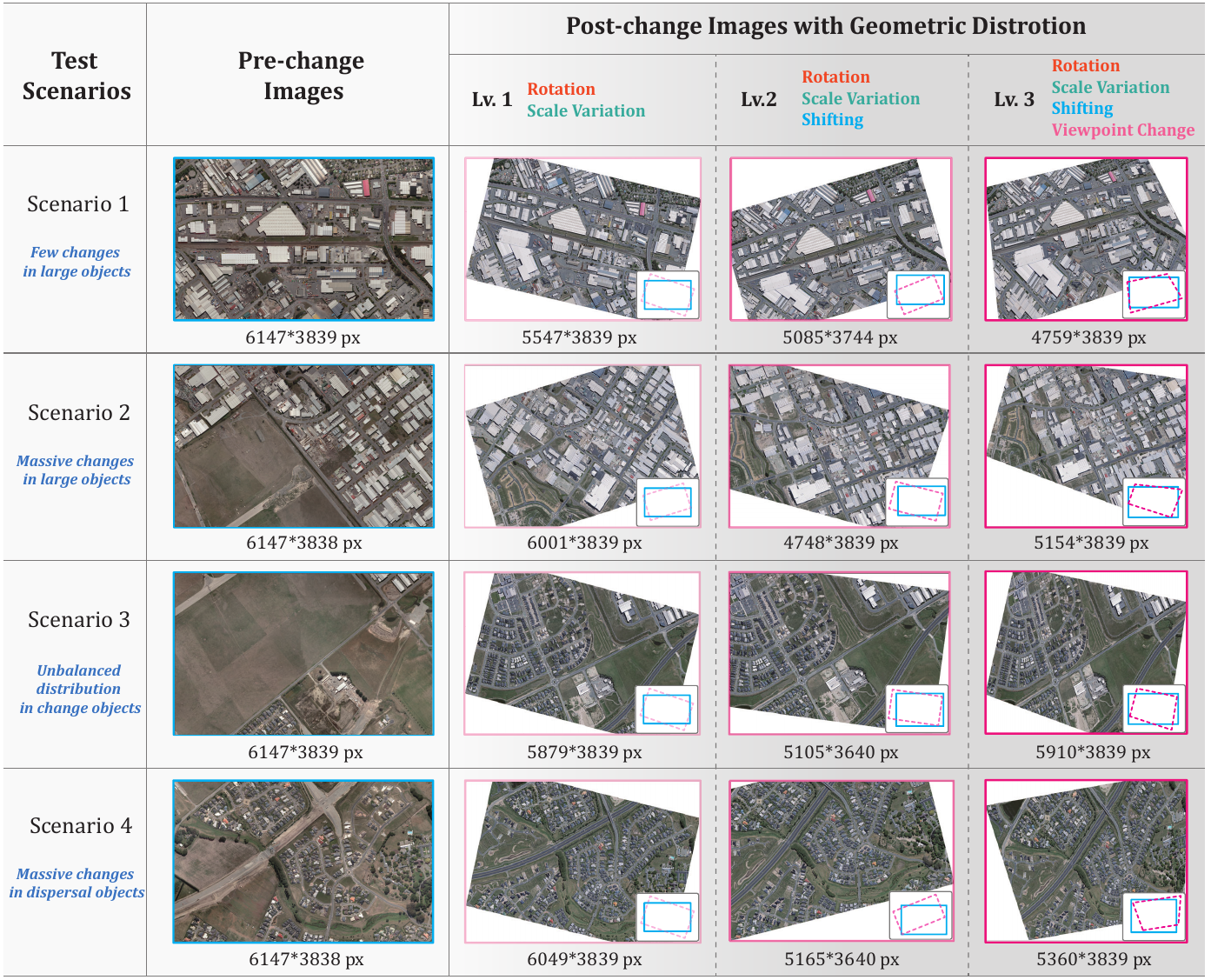}
    \caption{Illustration of the three-level geometric distortions in the proposed \textit{WarpCD} dataset. }
    \label{fig:warpcd}
\end{figure*}

\begin{itemize}
    \item Image Registration Scenarios: The \textit{WHU-CD} dataset \cite{ji2019full} is used to create the benchmark for image registration task. Specifically, we select 4 basic regions of $6147 * 3839 $ pixels, where 3 levels of geometric distortions are applied to create a new dataset containing 12 scenarios with different matching difficulties. For the spatial rotation transformation, we apply random rotation with angles ranging from 0 to 30 degrees.  For the shifting factor, the pixel offset within 20\% of the image size is implemented on random directions. The created registration dataset ``\textit{WarpCD}'' are shown in Fig \ref{fig:warpcd}.

    \item Change Detection Datasets: Note that since few research has included the distortion scenarios in CD task, we particularly test the proposed ``MatchCD'' framework on the ``\textit{WarpCD}' dataset. 
    For further evaluation, the remaining portion of ``\textit{WHU-CD}'' dataset excluding the registration regions is used as the training set for the selected CD methods.
\end{itemize}

\subsection{Evaluation Metrics}
To evaluating the effectiveness of the proposed MatchCD framework, we select three main evaluation metrics: F1-Score, IoU and OA. The specific calculation for metrics is as follows:
\begin{small}
\begin{align}
    \mathit{F1} &= 2 \times Precision \times Recall / (Precision + Recall) \\ 
    \mathit{IoU} &= TP / (TP + FN + FP) \\
    \mathit{OA} &= (TP+TN)/(TP+FN+FP+TN)
\end{align}
\end{small}where, $TP, FP, TN$, and $FN$ are components in the confusion matrix, representing true positive, false positive, and false negative, respectively. $Precision$ can be acquired by $TP/(TP+FP)$, and $Recall$ is calculated by $TP/(TP+FN)$.

\subsection{Implementation Details}
\subsubsection{Selected Downstream Image Registration Methods}
Three well-known image registration methods \textit{SIFT} \cite{lowe2004distinctive}, LightGlue \cite{lindenberger2023lightglue} and GIM \cite{xuelun2024gim} are selected for comparison on the created \textit{WarpCD} datasets.

\subsubsection{Selected Downstream RS CD Methods}
Several RS CD methods are selected for comparison on the open-source datasets and the proposed ``\textit{WarpCD}'' dataset. 

\begin{itemize}
    \item FC-EF \cite{daudt2018fully}: Single branch CNN model, in which the early fusion strategy is applied to the input images. 
	
    \item FC-Siam-conc \cite{daudt2018fully}: uses a dual-branch CNN architecture, where each branch processes the single temporal image and incorporates feature fusion by long-range skip connections.
	
    \item FC-Siam-diff \cite{daudt2018fully}: A siamese model is utilized, and feature differential and fusion are applied during the upsampling stage.
	
    \item SNUNet-CD \cite{fang2021snunet}: adopts U-Net++ architecture for bi-temporal change detection. Attention mechanism is introduced to connect hierarchical features.
	
    \item USSFCNet \cite{lei2023ultralightweight}: utilizes a lightweight model design with spatial-spectral fusion to enrich the feature representation.

    \item SEIFNet \cite{huang2024spatiotemporal}: A spatiotemporal enhancement and inter-level fusion network is proposed for spatiotemporal difference extraction. In addition, a refinement module improve boundary details and internal consistency are proposed for CD task.
\end{itemize}

\subsubsection{Model Training and Testing}
The training process is implemented on an NVIDIA RTX 3090 GPU. In the pre-training stage, the MatchCD framework is trained for 100 epochs with a batch size of 64. The parameters are optimized by AdamW optimizer and the initial learning rate is set to 0.005 with cosine decay. 

In the downstream stage, the selected CD methods are optimized by the AdamW optimizer with a momentum of 0.999 and weight decay of 0.01. The batch size for fine-tuning is set to 8. For the image registration evaluation, we directly test the proposed method with some commonly-used image registration methods with visual comparison. Then the registered images are segmented into patches and sent to the downstream CD task.

\subsection{Experimental Results}

\subsubsection{Training-free Image Registration}
\label{registration}

\begin{figure}[!hbp]
    \centering
    \scalebox{1.0}{
    \includegraphics[width=\linewidth]{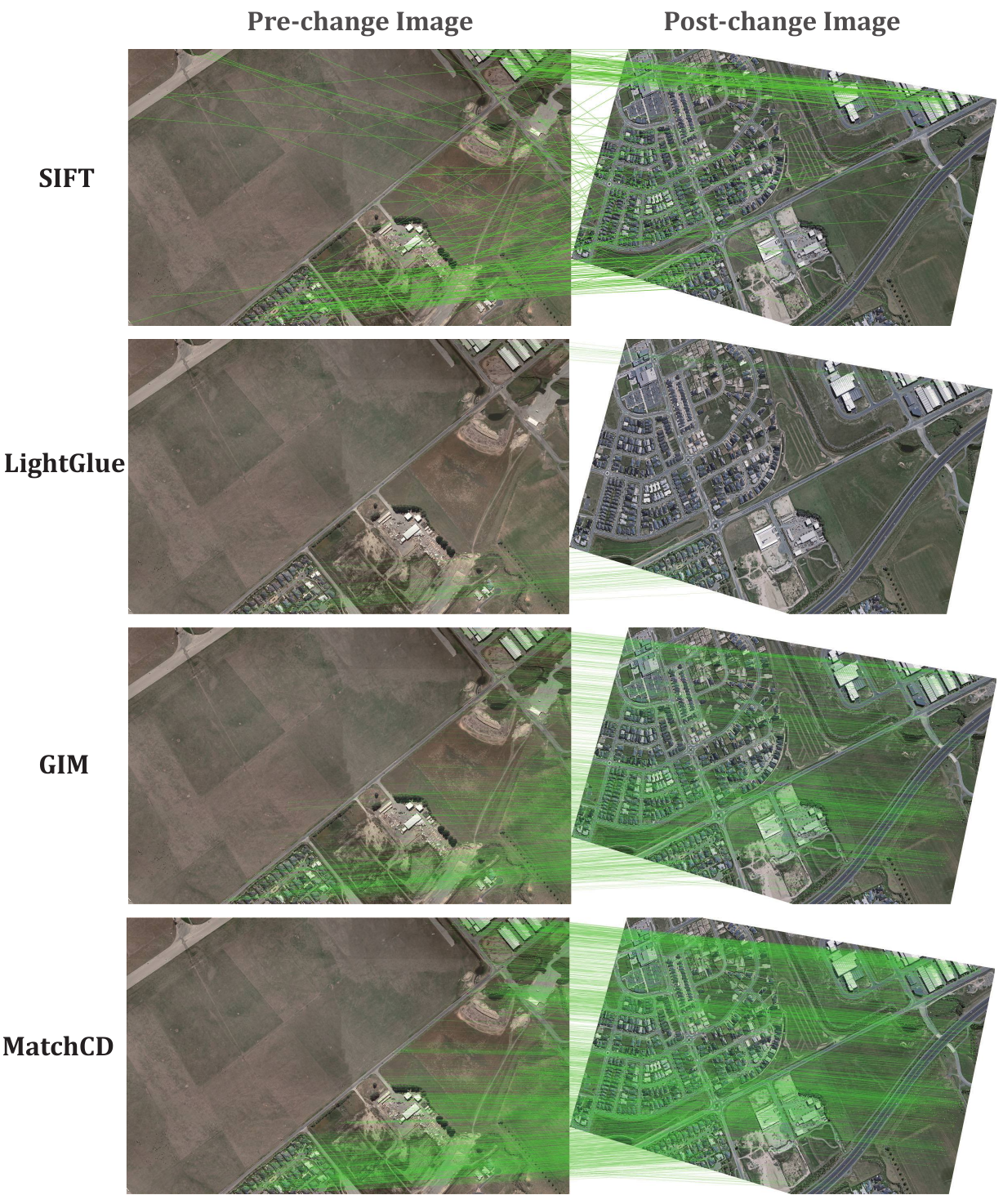}
    }
    \caption{Comparative performance of image matching algorithms on \textit{WarpCD} dataset (Lv. 3 difficulty). }
    \label{fig:matching_comparison}
\end{figure}

\begin{figure*}[!htp]
    \centering
    \includegraphics[width=0.95\linewidth]{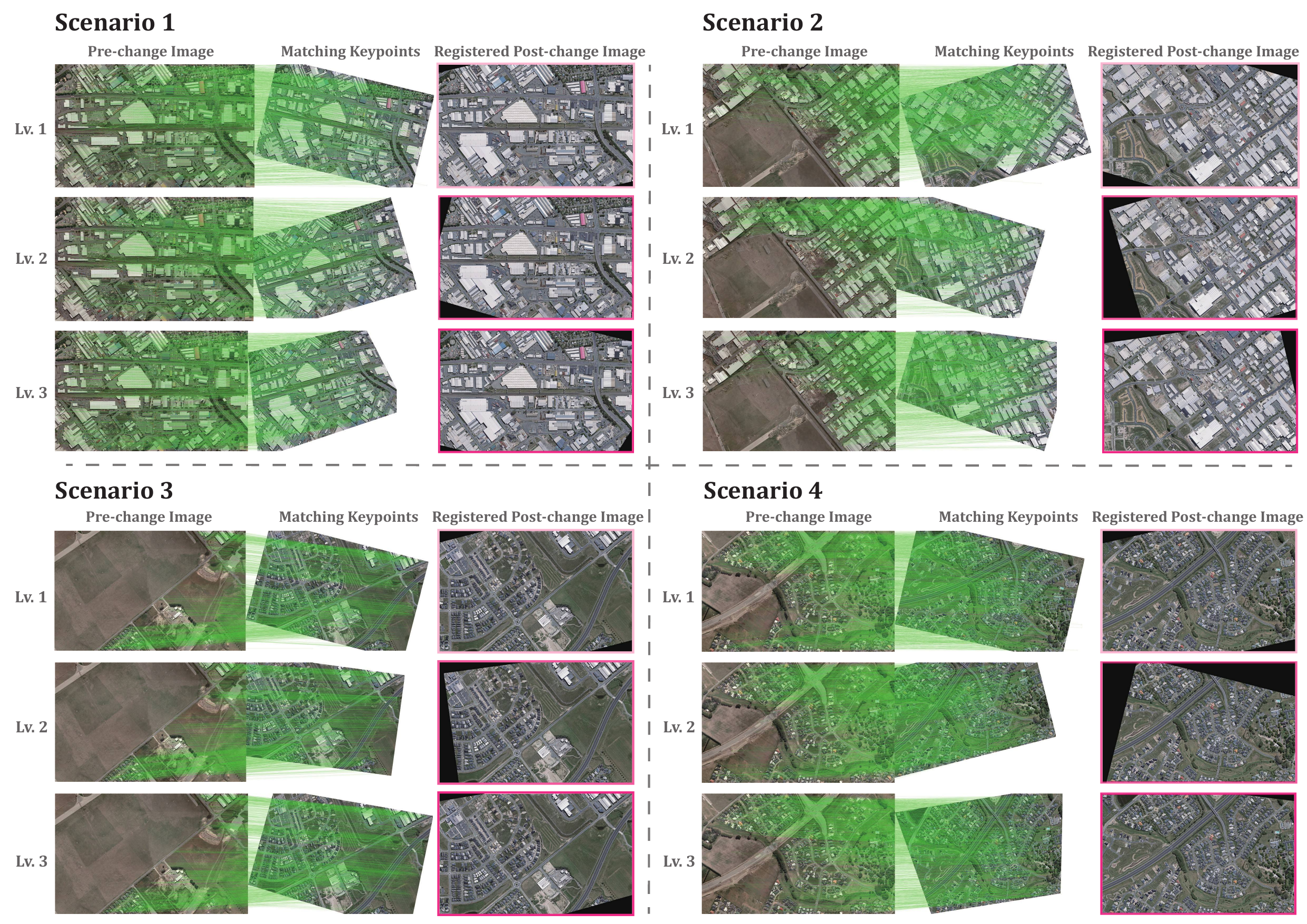}
    \caption{Matching performance of the proposed MatchCD method in all distorted test scenarios in \textit{WarpCD} dataset. }
    \label{fig:matchcd_matching}
\end{figure*}

As the prior task in full-scenario ``Registration-Detection'', we first apply the proposed MatchCD method in the \textit{WarpCD} dataset. The effectiveness of the proposed method could be clearly demonstrated through the visual comparison. In addition, the registered images serve as the test scenario for the CD baselines. 

The specific image registration results of MatchCD method is shown in Fig. \ref{fig:matchcd_matching}. The first and second columns in each scenario are the pre-change and post-change images. The matching keypoints of MatchCD are drawn across the two images. The registered post-change image calibrated by MatchCD method is shown in the last column. Along the changing level from $Lv.1$ to $Lv.3$, significant geometric distortions such as scale variation and view-point change are gradually added to the post-change images to stimulate the imaging condition of different sensors in real-world scenarios. We observe that although the scenarios in \textit{WarpCD} dataset is quite complex with multiple ground objects as well as changes caused by seasonal and manual factors, the MatchCD method can still achieve satisfying performance in all test scenarios.

It is worth noting that massive changes have occurred between the test image pairs of $Scenario$ \textit{3} in the \textit{WarpCD} dataset, which is a great challenge to feature matching. In this case, the proposed MatchCD method exploits the unchanged objects in the bottom and top right corner as the matching reference. The hierarchical matching strategy using deep feature maps compensates for the lack of keypoints detected on the original images.

\begin{table}[!htbp]
    \centering
    \caption{Comparison between the selected image registration methods.}
    \renewcommand{\arraystretch}{1.7}
    \setlength{\tabcolsep}{6pt}
    \footnotesize
    \begin{tabular}{c|c|c|c}
        \Xhline{1pt}
        \textbf{Method} & \textbf{Descriptor} & \textbf{Paradigm} & \textbf{Keypoint number} 
         \\ \hline
         SIFT & Manually designed  & Training-free & 982
         \\ \hline
         LightGlue & Learning-based  & Sup. Training & 374
         \\ \hline
         GIM & Learning-based  & Sup. Training & 4982
         \\ \hline
         \rowcolor{gray!20}
         \textbf{MatchCD} & \textbf{Learning-based}   & \textbf{Training-free} & \textbf{14867}
         \\ \Xhline{1pt}
    \end{tabular}
    \label{tab:matching_method}
\end{table}

For the purpose of further investigating the image registration performance of MatchCD, we consider this scenario as a hard sample and perform additional evaluations with the selected image registration methods. The experimental results are illustrated in Fig. \ref{fig:matching_comparison}. As a commonly used  matching algorithm, SIFT fails to match in this scenario with sparse objects. The LightGlue method only acquire a few matching keypoints which is adequate for the subsequent calibration procedure. 

As shown in Table \ref{tab:matching_method}, the GIM matcher extracts relative more keypoints around the buildings than the above two methods. Benefiting from our improved matching strategy applied to the vanilla GIM matcher, the MatchCD method obtains numerous and uniformly distributed keypoints.

\subsubsection{Quantitative Analysis for Change Detection}
\label{quantitative}
In this section, comparative experiments are performed on the selected CD methods with several datasets to evaluate the change detection performance. 

\textbf{(a) Evaluation on \textit{WarpCD} Dataset}

\textbf{Input:} The \textit{unregistered} image pairs in \textit{WarpCD} dataset.

\begin{table}[!htbp]
    \centering
    \caption{Change detection results of MatchCD method on WarpCD dataset.}
    \renewcommand{\arraystretch}{1.8}
    \setlength{\tabcolsep}{7pt}
    \footnotesize
    \begin{tabular}{c|c|c|c|c}
        \Xhline{1pt}
        \textbf{Method} & \textbf{Dataset} & \textbf{F1-Score} & \textbf{IoU} & \textbf{OA}
         \\ \hline
         \multirow{4}{*}{MatchCD} & \textit{Distorted Scenario 1} & \textbf{71.85} & \textbf{56.07} & \textbf{96.64}
         \\ \cline{2-5}
         
          & \textit{Distorted Scenario 2} & \textbf{84.80} & \textbf{73.61} & \textbf{97.78}
         \\ \cline{2-5}

          & \textit{Distorted Scenario 3} & \textbf{74.13} & \textbf{58.90} & \textbf{94.97}
         \\ \cline{2-5}
         
         & \textit{Distorted Scenario 4} & \textbf{80.60} & \textbf{67.51} & \textbf{98.37}
         \\ \Xhline{1pt}
    \end{tabular}
    \label{tab:matchcd_warpcd}
\end{table}

We first apply the MatchCD method on the proposed \textit{WarpCD} dataset to demonstrate its capability of simultaneous image registration and change detection. Since none of the other CD methods take the geometric distortion factors into consideration, only the proposed MatchCD method is evaluated in this case. 

As shown in Table \ref{tab:matchcd_warpcd}, although large geometric distortions exist in \textit{WarpCD} dataset, the proposed MatchCD method is still able to overcome the unalignment issue and obtain reliable performance in all scenarios. We notice the metrics of MatchCD in some test scenarios are relatively lower than that in the aligned scenarios, which is subsequently listed in Table \ref{tab:s1_warpcd} to Table \ref{tab:s4_warpcd}. The reason for such decrease is that the distorted $T_2$ images are firstly registered and cropped by the extracted overlap boundaries in this situation. The test regions will be smaller than the standard aligned scenarios, leading to the decent in the evaluation metrics. 

\textbf{(b) Evaluation on Registered \textit{WarpCD} Dataset}

\textbf{Input:} The \textit{aligned} image pairs of \textit{WarpCD} dataset. 

\begin{table}[!htp]
    \centering
    \caption{Evaluation performance in the Scenario 1 of \textit{WarpCD} dataset.}
    \renewcommand{\arraystretch}{1.75}
    \setlength{\tabcolsep}{7pt}
    \footnotesize
    \begin{tabular}{c|c|c c c}
        \Xhline{1pt}
        \textbf{Dataset} & \textbf{Method} & \textbf{F1-Score} & \textbf{IoU} & \textbf{OA}
         \\ \hline
         \multirow{8}{*}{\textit{Aligned Scenario 1}} & FC-EF & \cellcolor{skyblue!50}70.12 & \cellcolor{skyblue!75}53.99 & \cellcolor{skyblue!75}96.52
         \\ \cline{2-2}
         & FC-Siam-Conc & \cellcolor{skyblue!20}61.87 & \cellcolor{skyblue!20}44.79 & \cellcolor{skyblue!20}93.31
         \\ \cline{2-2}
         & FC-Siam-Diff & \cellcolor{skyblue!20}59.57 & \cellcolor{skyblue!20}42.42 & \cellcolor{skyblue!20}93.90
         \\ \cline{2-2}
         & SNUNet-CD & \cellcolor{skyblue!50}64.28 & \cellcolor{skyblue!50}47.36 & \cellcolor{skyblue!50}96.22
         \\ \cline{2-2}
         & USSFCNet & \cellcolor{skyblue!30}56.39 & \cellcolor{skyblue!20}39.26 & \cellcolor{skyblue!50}95.81
         \\ \cline{2-2}
         & SEIFNet & \cellcolor{skyblue!50}70.42 & \cellcolor{skyblue!50}54.34 & \cellcolor{skyblue!50}95.84
         \\ \cline{2-2}
          & \textbf{MatchCD-base} & \cellcolor{skyblue!100}\textbf{75.87} & \cellcolor{skyblue!100}\textbf{61.12} & \cellcolor{skyblue!100}\textbf{97.10}
          \\ \cline{2-2}
         & \textbf{MatchCD} & \cellcolor{skyblue!75}\textbf{71.23} & \cellcolor{skyblue!75}\textbf{55.31} & \cellcolor{skyblue!75}\textbf{96.76}
         \\ \Xhline{1pt}
    \end{tabular}
    \label{tab:s1_warpcd}
\end{table}

\begin{table}[!htbp]
    \centering
    \caption{Evaluation performance in the Scenario 2 of \textit{WarpCD} dataset.}
    \renewcommand{\arraystretch}{1.75}
    \setlength{\tabcolsep}{7pt}
    \footnotesize
    \begin{tabular}{c|c|c c c}
        \Xhline{1pt}
        \textbf{Dataset} & \textbf{Method} & \textbf{F1-Score} & \textbf{IoU} & \textbf{OA}
         \\ \hline
         \multirow{8}{*}{\textit{Aligned Scenario 2}} & FC-EF & \cellcolor{grassgreen!50}81.04 & \cellcolor{grassgreen!50}68.12 & \cellcolor{grassgreen!50}97.07
         \\ \cline{2-2}
         & FC-Siam-Conc & \cellcolor{grassgreen!20}62.98 & \cellcolor{grassgreen!20}45.97 & \cellcolor{grassgreen!20}92.17
         \\ \cline{2-2}
         & FC-Siam-Diff & \cellcolor{grassgreen!20}63.40 & \cellcolor{grassgreen!20}46.41 & \cellcolor{grassgreen!20}92.40
         \\ \cline{2-2}
         & SNUNet-CD & \cellcolor{grassgreen!50}84.20 & \cellcolor{grassgreen!50}72.71 & \cellcolor{grassgreen!75}98.04
         \\ \cline{2-2}
         & USSFCNet & \cellcolor{grassgreen!30}71.73 & \cellcolor{grassgreen!20}55.92 & \cellcolor{grassgreen!50}96.87
         \\ \cline{2-2}
         & SEIFNet & \cellcolor{grassgreen!75}85.10 & \cellcolor{grassgreen!75}74.06 & \cellcolor{grassgreen!50}97.72
         \\ \cline{2-2}
          & \textbf{MatchCD-base} & \cellcolor{grassgreen!50}\textbf{83.49} & \cellcolor{grassgreen!50}\textbf{71.67} & \cellcolor{grassgreen!75}\textbf{97.50}
          \\ \cline{2-2}
         & \textbf{MatchCD} & \cellcolor{grassgreen!100}\textbf{87.01} & \cellcolor{grassgreen!100}\textbf{77.02} & \cellcolor{grassgreen!100}\textbf{98.15}
         \\ \Xhline{1pt}
    \end{tabular}
    \label{tab:s2_warpcd}
\end{table}

\begin{table}[!htbp]
    \centering
    \caption{Evaluation performance in the Scenario 3 of \textit{WarpCD} dataset.}
    \renewcommand{\arraystretch}{1.75}
    \setlength{\tabcolsep}{7pt}
    \footnotesize
    \begin{tabular}{c|c|c c c}
        \Xhline{1pt}
        \textbf{Dataset} & \textbf{Method} & \textbf{F1-Score} & \textbf{IoU} & \textbf{OA}
         \\ \hline
         \multirow{8}{*}{\textit{Aligned Scenario 3}} & FC-EF & \cellcolor{orange!50}85.55 & \cellcolor{orange!50}74.75 & \cellcolor{orange!30}97.04
         \\ \cline{2-2}
         & FC-Siam-Conc & \cellcolor{orange!20}78.80 & \cellcolor{orange!20}65.01 & \cellcolor{orange!20}95.08
         \\ \cline{2-2}
         & FC-Siam-Diff & \cellcolor{orange!20}70.88 & \cellcolor{orange!20}54.89 & \cellcolor{orange!20}92.56
         \\ \cline{2-2}
         & SNUNet-CD & \cellcolor{orange!75}86.07 & \cellcolor{orange!75}75.54 & \cellcolor{orange!75}97.51
         \\ \cline{2-2}
         & USSFCNet & \cellcolor{orange!30}85.28 & \cellcolor{orange!20}74.34 & \cellcolor{orange!75}97.36
         \\ \cline{2-2}
         & SEIFNet & \cellcolor{orange!50}85.38 & \cellcolor{orange!50}74.49 & \cellcolor{orange!50}96.91
         \\ \cline{2-2}
          & \textbf{MatchCD-base} & \cellcolor{orange!75}\textbf{85.86} & \cellcolor{orange!75}\textbf{75.23} & \cellcolor{orange!50}\textbf{97.14}
          \\ \cline{2-2}
         & \textbf{MatchCD} & \cellcolor{orange!100}\textbf{89.16} & \cellcolor{orange!100}\textbf{80.45} & \cellcolor{orange!100}\textbf{97.90}
         \\ \Xhline{1pt}
    \end{tabular}
    \label{tab:s3_warpcd}
\end{table}

\begin{table}[!htbp]
    \centering
    \caption{Evaluation performance in the Scenario 4 of \textit{WarpCD} dataset.}
    \label{tab:s4_warpcd}
    \renewcommand{\arraystretch}{1.75}
    \setlength{\tabcolsep}{7pt}
    \footnotesize
    \begin{tabular}{c|c|c c c}
         \Xhline{1pt}
         \textbf{Dataset} & \textbf{Method} & \textbf{F1-Score} & \textbf{IoU} & \textbf{OA}
         \\ \hline
         \multirow{8}{*}{\textit{Aligned Scenario 4}} & FC-EF & \cellcolor{gold!75}83.74 & \cellcolor{gold!75}72.01 & \cellcolor{gold!100}98.58
         \\ \cline{2-2}
         & FC-Siam-Conc & \cellcolor{gold!30}73.63 & \cellcolor{gold!30}58.26 & \cellcolor{gold!30}97.25
         \\ \cline{2-2}
         & FC-Siam-Diff & \cellcolor{gold!20}63.37 & \cellcolor{gold!20}46.39 & \cellcolor{gold!20}95.73
         \\ \cline{2-2}
         & SNUNet-CD & \cellcolor{gold!75}82.96 & \cellcolor{gold!50}70.90 & \cellcolor{gold!75}98.49
         \\ \cline{2-2}
         & USSFCNet & \cellcolor{gold!50}74.31 & \cellcolor{gold!30}59.11 & \cellcolor{gold!50}98.09
         \\ \cline{2-2}
         & SEIFNet & \cellcolor{gold!50}73.52 & \cellcolor{gold!50}58.12 & \cellcolor{gold!30}97.14
         \\ \cline{2-2}
          & \textbf{MatchCD-base} & \cellcolor{gold!50}\textbf{81.92} & \cellcolor{gold!50}\textbf{69.38} & \cellcolor{gold!50}\textbf{98.37}
          \\ \cline{2-2}
         & \textbf{MatchCD} & \cellcolor{gold!100}\textbf{84.04} & \cellcolor{gold!100}\textbf{72.48} & \cellcolor{gold!100}\textbf{98.57}
         \\ \Xhline{1pt}
    \end{tabular}
\end{table}

Afterwards, we use the accurately registered images of \textit{WarpCD} dataset to generate image patch sets as standard benchmarks to perform fair comparison. Specifically, the CD module in the proposed MatchCD method is evaluated with the other selected CD methods. The comparative results are shown in Table \ref{tab:s1_warpcd} to Table \ref{tab:s4_warpcd}. The penultimate rows ``MatchCD-base'' in the tables represent that the MatchCD method is directly pre-trained on the unlabeled RS dataset without the instance-level contrastive optimization. 

Compared with the other selected CD methods, the proposed MatchCD achieves the highest performance in all the test scenarios with the superiority of 5.75\%/7.13\%, 1.97\%/2.96\%, 3.09\%/4.91\%, 0.30\%/0.47\% on F1-Score and IoU metrics, respectively. Benefiting from the prior knowledge from the pre-trained encoder and visual foundation model, the MatchCD method shows robust adaptability to the object change. In all comparative results, we notice a relative high performance obtained by FC-EF method. Delving into the testing scenarios, we are convinced that due to the imbalance of changed/unchanged samples, the FC-EF method is overfitting and biased towards an unchanged prediction, which leads to the increase in metrics.

As to the influence of MatchCD components, the metric elevation between MatchCD-base and MatchCD demonstrates that the instance contrastive optimization could stability enhance the performance in the majority of the scenarios. Further experiments of the MatchCD architecture are illustrated in the subsequent discussions of ablation sections. 

\textbf{(c) Evaluation on \textit{WHU-CD} Dataset}

In addition, a further investigation is performed on the remaining regions of \textit{WHU-CD} public dataset to prove the effectiveness of our proposed method. The specific results are shown in Table \ref{tab:whucd}. In contrast to the aforementioned datasets, \textit{WHU-CD} dataset possesses a relatively larger data volume which is more objective and challenging. It can be observed that our proposed MatchCD method outperforms the suboptimal algorithm by $0.16\%$ and $0.25\%$ in two main metrics of F1-Score and IoU, respectively.

\begin{table}[!hbp]
    \centering
    \caption{Evaluation performance in the WHU-CD dataset .}
    \renewcommand{\arraystretch}{1.75}
    \setlength{\tabcolsep}{9.5pt}
    \footnotesize
    \begin{tabular}{c|c|c c c}
        \Xhline{1pt}
        \textbf{Dataset} & \textbf{Method} & \textbf{F1-Score} & \textbf{IoU} & \textbf{OA}
         \\ \hline
         \multirow{7}{*}{\textit{WHU-CD}} & FC-EF & \cellcolor{gray!30}84.97 & \cellcolor{gray!30}73.87 & \cellcolor{gray!30}99.24
         \\ \cline{2-2}
         & FC-Siam-Conc & \cellcolor{gray!10}75.88 & \cellcolor{gray!10}61.13 & \cellcolor{gray!10}98.58
         \\ \cline{2-2}
         & FC-Siam-Diff & \cellcolor{gray!10}74.52 & \cellcolor{gray!10}59.39 & \cellcolor{gray!20}98.51
         \\ \cline{2-2}
         & SNUNet-CD & \cellcolor{gray!25}83.85 & \cellcolor{gray!25}72.19 & \cellcolor{gray!25}99.20
         \\ \cline{2-2}
         & USSFCNet & \cellcolor{gray!10}76.70 & \cellcolor{gray!10}62.20 & \cellcolor{gray!10}98.96
         \\ \cline{2-2}
         & SEIFNet & \cellcolor{gray!25}83.53 & \cellcolor{gray!30}71.72 & \cellcolor{gray!25}99.11
         \\ \cline{2-2}
         & \textbf{MatchCD} & \cellcolor{gray!50}\textbf{85.13} & \cellcolor{gray!50}\textbf{74.12} & \cellcolor{gray!50}\textbf{99.25}
         \\ \Xhline{1pt}
    \end{tabular}
    \label{tab:whucd}
\end{table}

In summary, extensive quantitative results demonstrate that, motivated by the generalizable pre-training framework, the proposed MatchCD framework possesses the capability of robust image registration and change detection in the real-world scenarios. Compared to the conventional fully-supervised paradigm, our MatchCD framework can be directly established on the unlabeled dataset without annotation. In addition, the training-free matching strategy utilized in MatchCD directly performs hierarchical matching based on the pre-trained weight, eliminating the requirement for manual matching in traditional workflows. 

\begin{figure*}[!htp]
    \centering
    \includegraphics[width=\linewidth]{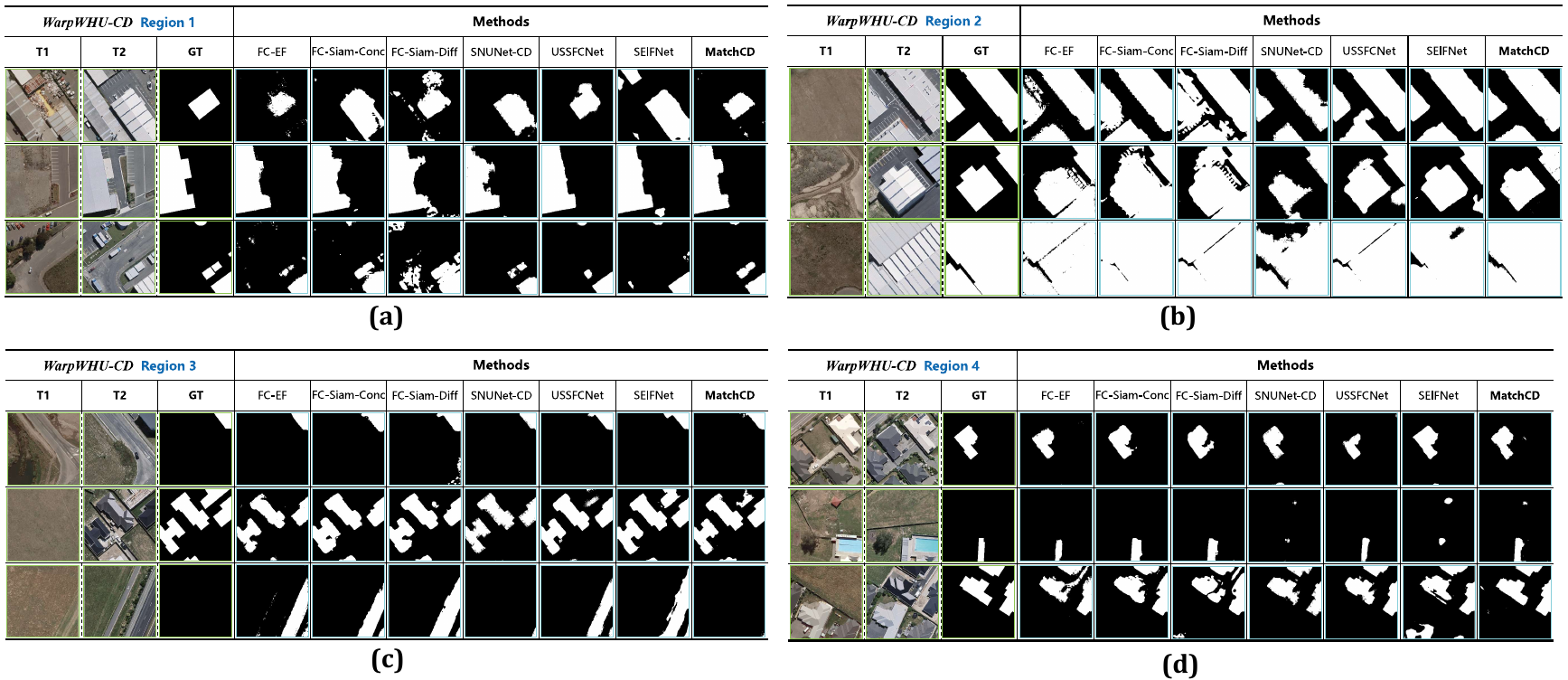}
    \caption{Illustration of detailed change detection results of the proposed MatchCD method in all the distorted test scenarios on the registered \textit{WarpCD} dataset. }
    \label{fig:visual_local}
\end{figure*}

\begin{figure*}[!htbp]
    \centering
    \includegraphics[width=\linewidth]{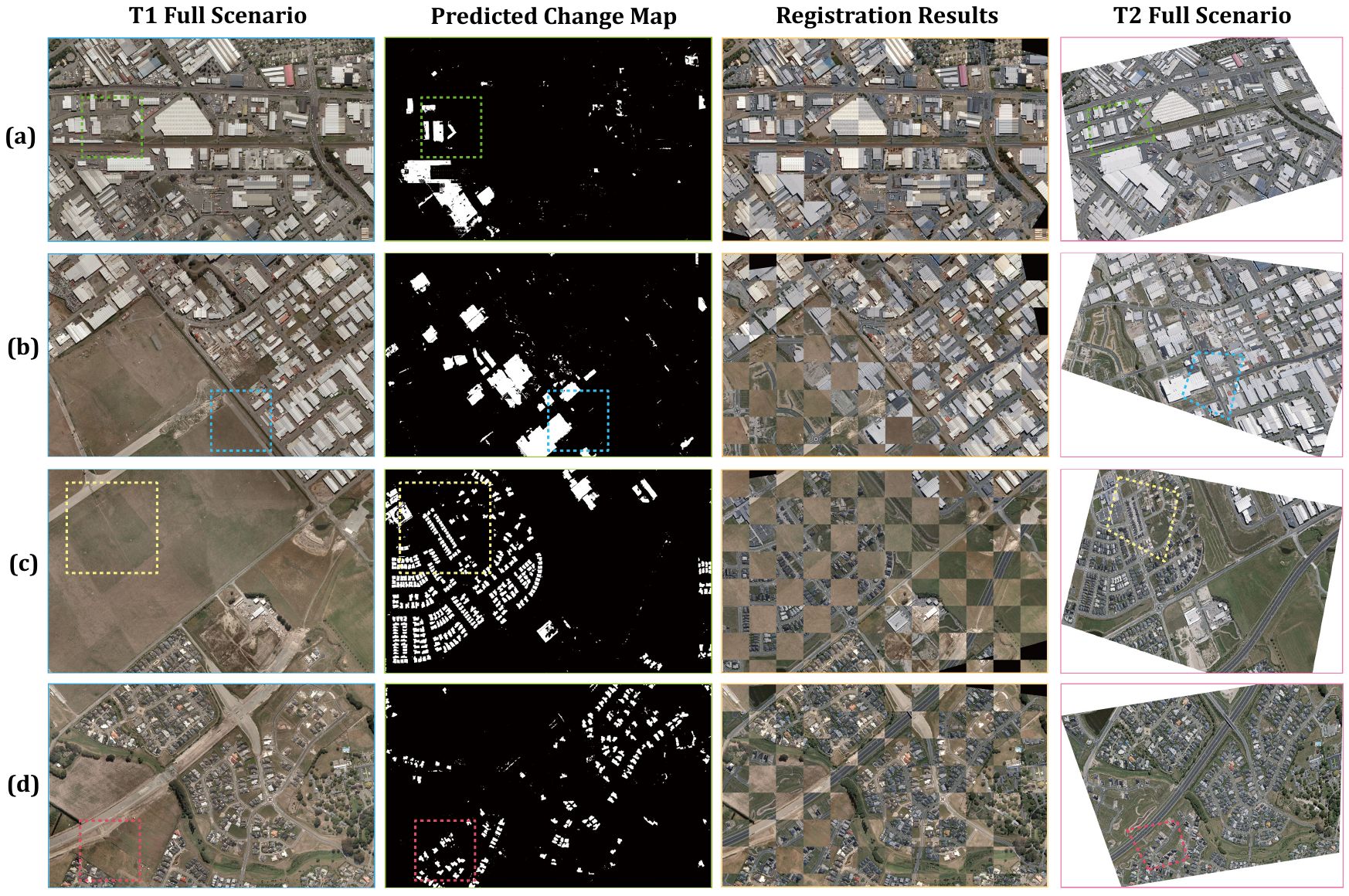}
    \caption{Illustration of full scenario change detection results of the proposed MatchCD method in all distorted test scenarios in \textit{WarpCD} dataset. }
    \label{fig:visual_global}
\end{figure*}

\subsubsection{Visualization Analysis for Change Detection}
\label{visual}

For further illustrating the effectiveness of the proposed MatchCD method, we visualize the prediction results for the selected CD methods on each scenario in \textit{WarpCD} dataset. The detailed comparison of visualization results are illustrated in Fig. \ref{fig:visual_local}. Through the horizontal comparison, we observe that our MatchCD method is capable of obtaining finer-grained details and precise geometric boundaries. We believe that benefiting from the guidance of prior knowledge provided by the FastSAM, MatchCD pays more attention to the geometrical properties of the objects during the optimization process. In addition, we notice that the proposed MatchCD also presents promising understanding capability for the semantic features of test scenarios. As illustrated in Fig.\ref{fig:visual_local}-(c), the third pair of samples shows a constructed highway which is evolved from a country road. Considering the main target is the change of building objects, the MatchCD method classifies both the highway and country road into the background category. While the majority of the other comparative CD methods incorrectly attribute the change of road to the building class. 

In order to more intuitively discriminate the performance of our MatchCD method, we implement a full-scene registration and change detection for all of the Lv. 3 scenarios in the \textit{WarCD} dataset. Afterwards, the visualization results of the final change maps are presented in Fig. \ref{fig:visual_global}. In each scenario, the same detection regions are highlighted with dashed boxes of the same color. It is obvious that the detection regions in the $T_2$ scenarios are presented as irregular polygons due to the bi-temporal geometric distortions. The proposed MatchCD method is capable of fully leveraging the unchanged targets in the scenario as a reference for image calibration. Furthermore, the changed regions are projected into the coordinate system of the pre-change image at the pixel level. With the guidance of the pre-trained encoder, our method can overcome the pseudo-changes caused by illumination and seasonal factors between the pre/post-change images and extract accurate boundaries of the change objects.

\section{Discussions}
We provide a further investigation of the main components and configurations of MatchCD method in this section.
\label{discussion}

\subsection{Ablation Experiments}
\label{sec:ablation}
In this part, several ablation studies are established to evaluate the effectiveness of the main components in the proposed MatchCD framework. The related study contains three main experiments as follows. 

\subsubsection{Attention Mechanism}
In the implementation of the MatchCD framework, we utilize the attention mechanism to fuse the knowledge from different domains. Then the fused features are sent to the decoder to calculate the changed regions. To keep the whole framework concise, we select several simple attentions for comparison: self attention, cross attention and their sequential combination. The results demonstrate that cross attention provides more enhancement to the MatchCD framework, in which various features from multimodal encoders are incorporated to the common latent space.

\begin{figure}[!htp]
    \centering
    \includegraphics[width=0.9\linewidth]{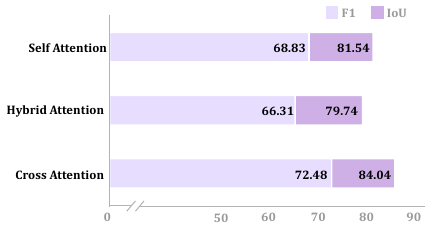}
    \caption{Comparison of different attention mechanisms applied in MatchCD for change detection.}
    \label{fig:att_ablation}
\end{figure}

\subsubsection{Number of Attention Heads} 
\begin{figure}[!htp]
    \centering
    \includegraphics[width=0.85\linewidth]{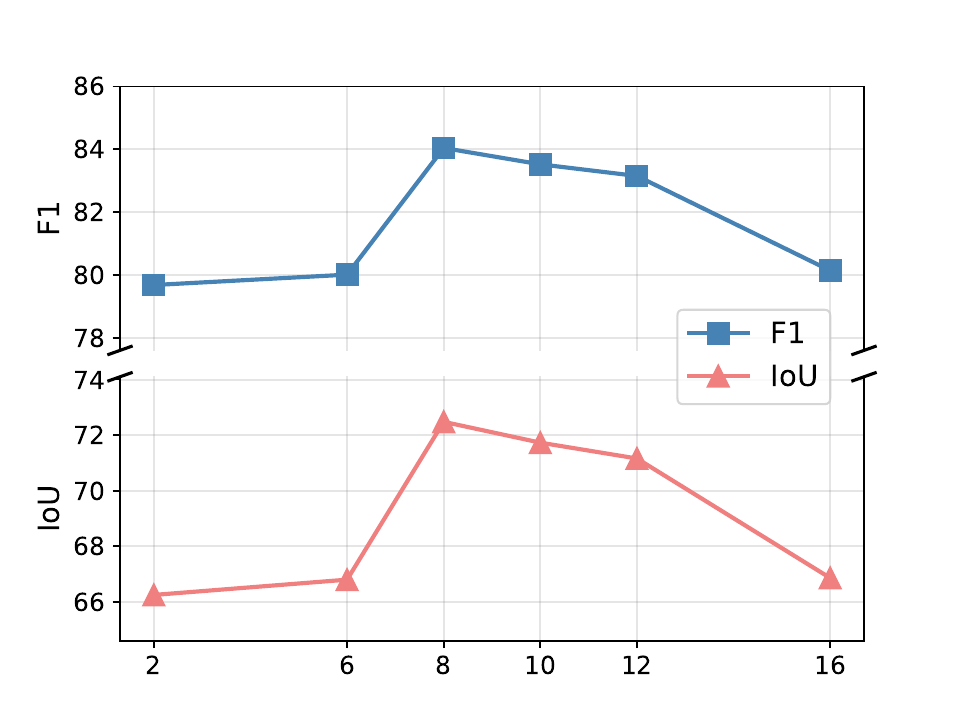}
    \caption{Ablation study of attention head number in MatchCD for change detection.}
    \label{fig:num_head_ablation}
\end{figure}

We also investigate the effect of the number of attention heads on the CD task. Specifically, we establish the MatchCD framework with different numbers of attention heads and fine-tune them on the \textit{WarpCD} dataset. From the results shown in Fig. \ref{fig:num_head_ablation}, a significant improvement is observed in the performance of both F1-Score and IoU metrics when the number of head is set to 8. It demonstrates that a moderate value of the selection of attention head number will be more instructive for down-stream CD task. 

\subsubsection{Model Components} For the main components of the MatchCD model, we select both the frozen pre-trained MatchCD encoder and FastSAM module for feature encoding, and the U-net++ decoder for the calculation of change map. Therefore, we explore the impact of these components on the downstream CD task. Through the comparison, we observe that the utilization of FastSAM prior knowledge is able to enhance the CD performance.
\begin{table}[!htp]
    \centering
    \caption{Ablation experiments of MatchCD encoder. (Pre. Enc. represents the pre-trained MatchCD encoder.) }
    \label{tab:ablation of encoder}
    \renewcommand{\arraystretch}{1.65}
    \setlength{\tabcolsep}{6.6pt}
    \footnotesize
    {
        \begin{tabular}{c|c|c|c|c}
            \Xhline{1pt}
            \textbf{Baseline} & \textbf{+ Pre. Enc.} & \textbf{+ FastSAM} & \textbf{F1} & \textbf{IoU}
            \\ \Xhline{0.8pt}
            \multirow{2}{*}{MatchCD} & \ding{51} & \ding{55} & 81.54 & 68.83
            \\ \cline{2-5}
             & \cellcolor[gray]{0.9} \ding{51} & \cellcolor[gray]{0.9} \ding{51}  & \cellcolor[gray]{0.9} 84.04 \textsubscript{\textcolor{darkgreen}{+2.50}} & \cellcolor[gray]{0.9} 72.48 \textsubscript{\textcolor{darkgreen}{+3.65}}
            \\ \Xhline{1pt}
        \end{tabular}
	}
\end{table}

Likewise, an ablation study of the decoder type is performed on basis of MatchCD framework. It demonstrates that the U-Net++ decoder provides an improved capability to project the tokens from attention layers to the output space. 
\begin{table}[!htp]
    \centering
    \caption{Ablation experiments of MatchCD decoder. }
    \label{tab:ablation of decoder}
    \renewcommand{\arraystretch}{1.65}
    \setlength{\tabcolsep}{6pt}
    \footnotesize
    {
        \begin{tabular}{c|c|c|c|c}
            \Xhline{1pt}
            \textbf{Baseline} & \textbf{+ ConvMLP} & \textbf{+ U-Net++} & \textbf{F1} & \textbf{IoU}
            \\ \Xhline{0.8pt}
            \multirow{2}{*}{MatchCD} & \ding{51} & \ding{55} & 70.14 & 54.01
            \\ \cline{2-5}
             & \cellcolor[gray]{0.9} \ding{55} & \cellcolor[gray]{0.9}\ding{51} & \cellcolor[gray]{0.9} 84.04 \textsubscript{\textcolor{darkgreen}{+13.90}} & \cellcolor[gray]{0.9}72.48 \textsubscript{\textcolor{darkgreen}{+18.47}}
            \\ \Xhline{1pt}
        \end{tabular}
	}
\end{table}

\subsection{Impact of Pre-training Stage}
In this part, we further discuss the impact of pre-training stage on the downstream tasks.

\subsubsection{Pre-training backbone} 
\begin{figure}[!htp]
    \centering
    \includegraphics[width=\linewidth]{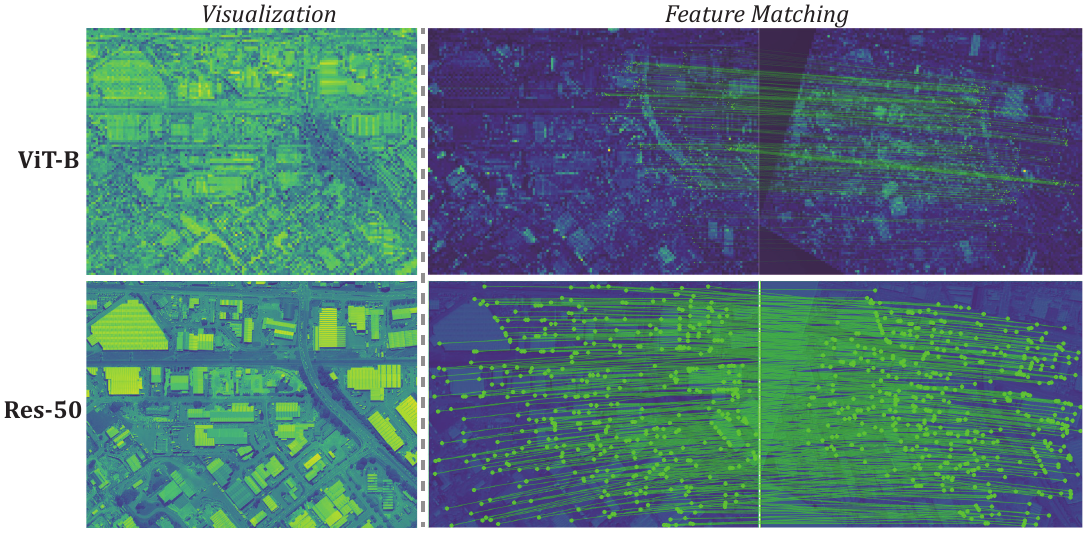}
    \caption{Visualization of latent features from different backbones.}
    \label{fig:pre_feature}
\end{figure}

We select two commonly used backbones: ResNet-50 and ViT-B as the MatchCD encoder for comparison, which is reused to provide robust features in both the downstream image registration and change detection tasks. The latent features are visualized in Fig. \ref{fig:pre_feature}. Due to the inherent patch partition operation, the resolution of latent feature from ViT-B is lower than that from ResNet-50 architecture. 

In addition, we perform feature matching between the bi-temporal latent features. As illustrated in Fig. \ref{fig:pre_feature}, benefiting from the distinctive features, the keypoints of ResNet-50 present a uniform distribution across the whole image, which facilitates the accurate estimation of homography matrix. In contrast, several mismatching can be observed on the matching results of ViT-B bi-temporal features. Considering the latent features are essential for hierarchical matching and subsequent change detection, the ResNet-50 architecture with fine-grained feature maps is selected as the MatchCD backbone. 

\subsubsection{Pre-training Paradigm} Besides the backbone architecture of pre-training stage, we also explore the effect of pre-training paradigm. Specifically, we select several self-supervised methods for pre-training, and then evaluate the performance in change detection task. Contrast to the proposed MatchCD method, SimSiam \cite{chen2021exploring} and DINO \cite{caron2021emerging} all perform image-level contrastive learning. Our MatchCD leverages the multiple objects in the scenario to establish instance-level contrastive optimization, prompting the backbone to capture the detailed patterns in downstream tasks.

\begin{figure}[!htp]
    \centering
    \scalebox{0.9}{
    \includegraphics[width=\linewidth]{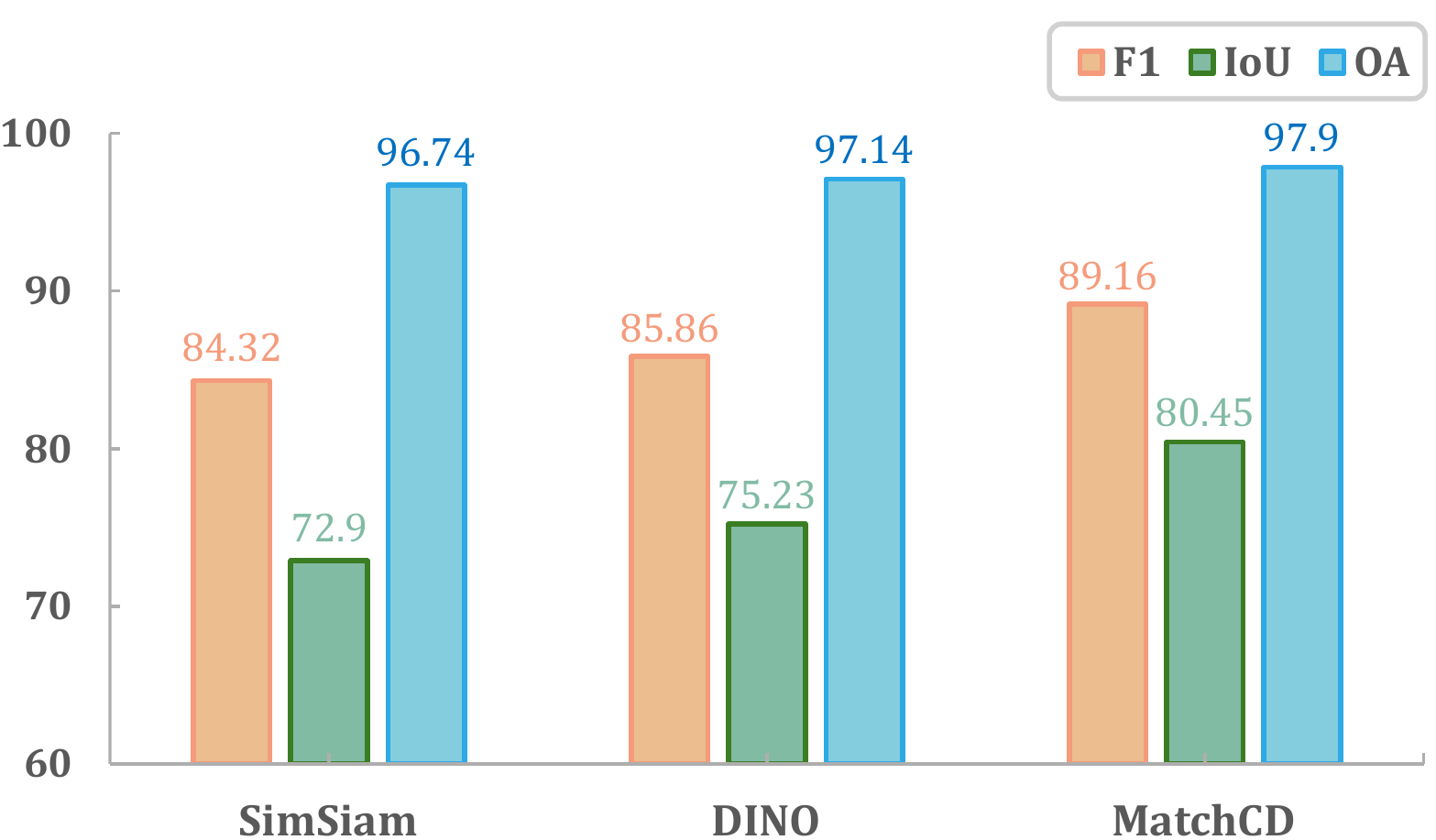}
    }
    \caption{Comparison of different self-supervised pre-training paradigms.}
    \label{fig:pre_paradigm}
\end{figure}

\subsection{Effect of Overlap Detection}
As shown in Fig. \ref{fig:change_detection}, the initial change map is cropped with the overlap boundary to generate the final change map. In this part, we quantify the implications of overlap detection for the downstream CD task. The comparative metrics are shown in Table. \ref{tab:overlap_detection}, reflecting that the overlap detection can largely eliminate the pseudo changes caused by the invalid regions after image registration. We notice a slight decrease of metrics in the scenario 3 dataset, which is due to the change in test region after the cropping operation.

\begin{table}[!htbp]
    \centering
    \caption{Comparative results for the effect of overlap detection.}
    \renewcommand{\arraystretch}{1.85}
    \setlength{\tabcolsep}{1.2pt}
    \footnotesize
    \begin{tabular}{c|cc|cc|cc|cc}
        \Xhline{1pt}
        \multicolumn{1}{c|}{\multirow{2}[0]{*}{\textbf{Overlap Detection}}} & \multicolumn{2}{c|}{\textbf{Scenario 1}} & \multicolumn{2}{c|}{\textbf{Scenario 2}} & \multicolumn{2}{c|}{\textbf{Scenario 3}} & \multicolumn{2}{c}{\textbf{Scenario 4}} 
        \\
        & \multicolumn{1}{c}{\textbf{F1}} & \multicolumn{1}{c|}{\textbf{IoU}} & \multicolumn{1}{c}{\textbf{F1}} & \multicolumn{1}{c|}{\textbf{IoU}} & \multicolumn{1}{c}{\textbf{F1}} & \multicolumn{1}{c|}{\textbf{IoU}} & \multicolumn{1}{c}{\textbf{F1}} & \multicolumn{1}{c}{\textbf{IoU}} 
        \\ \hline
        \ding{55} & 69.02 & 52.70 & 83.26 & 71.31 & 80.78 & 67.76 & 77.80 & 63.66
        \\ \cline{1-9}
         \rowcolor{gray!20}
        \ding{51} & 71.85 & 56.07 & 84.80 & 73.61 & 74.13 & 58.90 & 80.60 & 67.51
        \\ \Xhline{1pt}
    
    \end{tabular}%
  \label{tab:overlap_detection}%
\end{table}%

\subsection{Limitation Analysis}
In this research, a generalizable pre-training framework is proposed to simultaneously provide robust features to image registration and change detection tasks. The extensive experiments demonstrate the effectiveness of MatchCD framework. Nevertheless, the following existing limitations require further exploration, revealing the directions of future research.

\begin{itemize}
    \item \textit{Evaluation of Image Registration: } In the full-scene remote sensing image registration task involved in this work, visual assessment and the performance of downstream change detection are employed to indirectly  reflect the accuracy of the registration. However, such an approach cannot directly evaluate the performance of the registration task. Therefore, the related benchmark and criterion for the evaluation of RS image registration task with geometric distortion need to be further explored.

    \item \textit{The Capability of Visual Foundation Models: } During the model optimization of downstream change detection, the zero-shot capability of pre-trained visual foundation model is utilized to provide binary guidance to the attention layers and MatchCD decoder. Nevertheless, the fine-tuning for the visual foundation model for specific tasks is not established limited by available computational resources. The efficient adaption of foundation models for downstream is another priority direction for the subsequent research. 
\end{itemize}

\section{Conclusion}

In this paper, we propose a generalizable pre-training framework named MatchCD for simultaneous image registration and change detection tasks. Specifically, the class-agnostic instances are first generated with the zero-shot capability of FastSAM model. An instance-level contrastive optimization is established to motivate the encoder to capture and understand the fine-grained patterns in the scenario. Afterwards, a hierarchical geometric estimation stage is designed as a training-free paradigm. With the guidance of pre-trained MatchCD features, the homography matrix between pre-change and post-change image is estimated by the matcher without training. Finally, the aligned bi-temporal images are sent into the change detection stage, in which the latent features from pre-trained MatchCD encoder and binary prior knowledge from the FastSAM are both introduced to prompt the classifier to calculate the changed tokens. Comprehensive evaluation of bitemporal image registration and change detection on both created \textit{WarpCD} dataset and public dataset demonstrate the effectiveness of our MatchCD framework. In future research, the existing and potential limitations will be explored to expand the applicability of the proposed MatchCD framework. 

\small
\bibliographystyle{IEEEtranN}
\bibliography{manuscript}
\end{document}